\documentclass{article}

\PassOptionsToPackage{numbers, compress}{natbib}

\usepackage[preprint]{neurips_2021}

\usepackage[utf8]{inputenc} 
\usepackage[T1]{fontenc}    
\usepackage{hyperref}       
\usepackage{url}            
\usepackage{booktabs}       
\usepackage{amsfonts}       
\usepackage{nicefrac}       
\usepackage{microtype}      
\usepackage{xcolor}         
\usepackage[caption=false]{subfig}
\usepackage{graphicx}
\usepackage{amsmath}

\usepackage{multirow}
\usepackage{amssymb}
\usepackage{longtable}
\usepackage{tabu}
\usepackage{amsthm}

\newcommand{\syn}{$L_{syn}$}
\newcommand{\ssim}{$L_{SSIM}$}
\newcommand{\lone}{$L_{L1}$}
\newcommand{\lsmth}{$L_{smth}$}

\newtheorem{definition}{Definition}

\usepackage{longtable}
\usepackage{tabu}
\usepackage{lscape}
\usepackage{hhline}

\title{On the Sins of Image Synthesis Loss for Self-supervised Depth Estimation}

\author{%
    Zhaoshuo Li, Nathan Drenkow, Hao Ding, Andy S. Ding, Alexander Lu, \\
    \AND
    Francis X. Creighton, Russell H. Taylor, Mathias Unberath \\
    Johns Hopkins University
}

\begin{document}

\maketitle

\begin{abstract}
  Scene depth estimation from stereo and monocular imagery is critical for extracting 3D information for downstream tasks such as scene understanding. Recently, learning-based methods for depth estimation have received much attention due to their high performance and flexibility in hardware choice. However, collecting ground truth data for supervised training of these algorithms is costly or outright impossible. This circumstance suggests a need for alternative learning approaches that do not require corresponding depth measurements. Indeed, self-supervised learning of depth estimation provides an increasingly popular alternative. It is based on the idea that observed frames can be synthesized from neighboring frames if accurate depth of the scene is known -- or in this case, estimated. We show empirically that -- contrary to common belief -- improvements in image synthesis do not necessitate improvement in depth estimation. Rather, optimizing for image synthesis can result in diverging performance with respect to the main prediction objective -- depth. We attribute this diverging phenomenon to aleatoric uncertainties, which originate from data. Based on our experiments on four datasets (spanning street, indoor, and medical) and five architectures (monocular and stereo), we conclude that this diverging phenomenon is independent of the dataset domain and not mitigated by commonly used regularization techniques. To underscore the importance of this finding, we include a survey of methods which use image synthesis, totaling 127 papers over the last six years. This observed divergence has not been previously reported or studied in depth, suggesting room for future improvement of self-supervised approaches which might be impacted the finding.
\end{abstract}

\section{Introduction}
Depth estimation from images has long been recognized as an important research area 
and has gained substantial interest in recent years given its utility for downstream tasks such as scene understanding, registration, navigation, and control. Learning-based models often work well under strong supervision, but acquiring the necessary ground truth depth data can be cost prohibitive or impossible and often requires additional planning and extra hardware \citep{godard2017unsupervised}. 

\begin{figure}[t]
    \centering
    \includegraphics[width=0.75\linewidth]{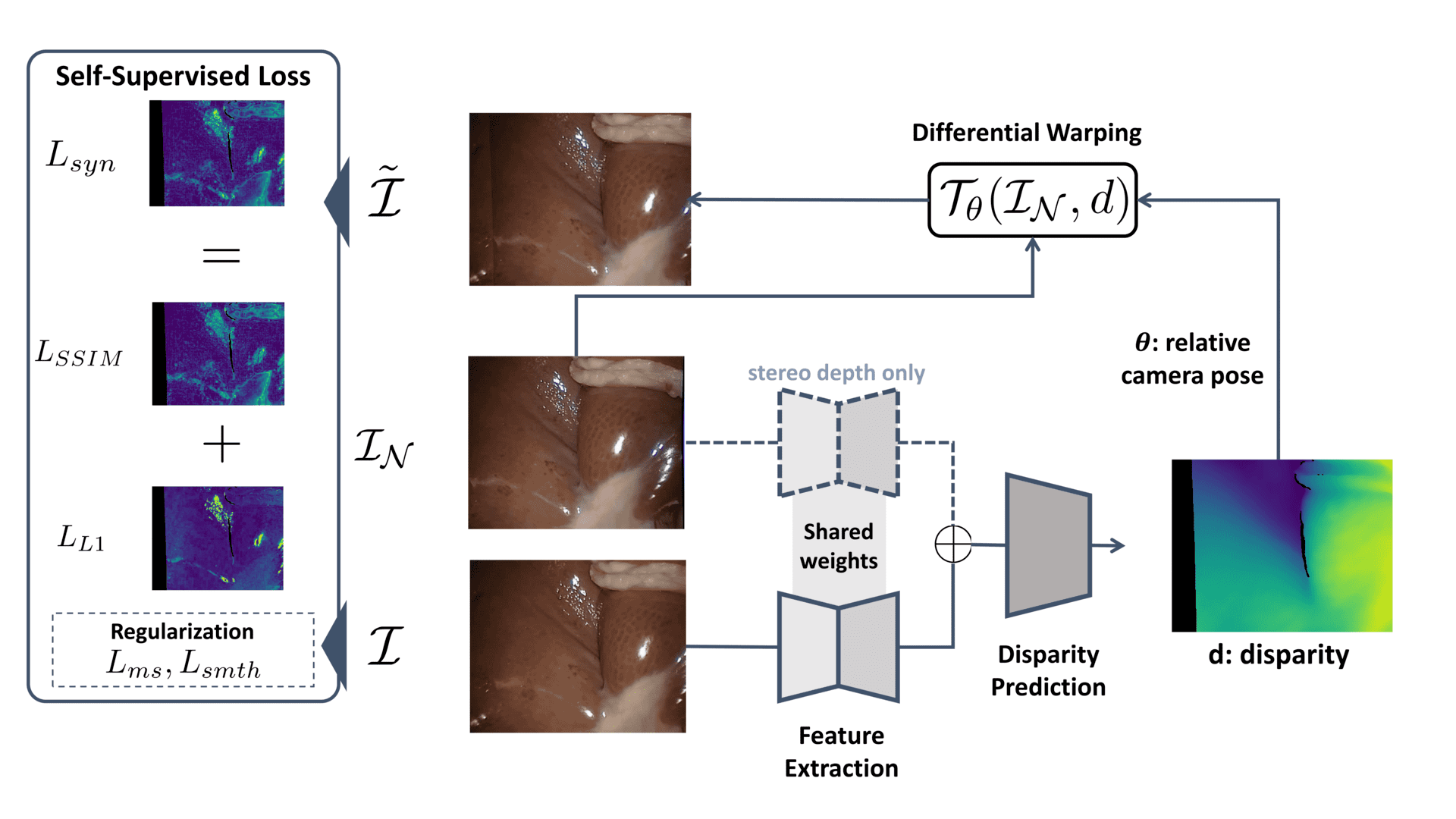}
    \caption{Overall process for self-supervised depth estimation using image synthesis. $\mathcal{I}$ is the image of the primary frame. $\mathcal{I}_\mathcal{N}$ is the neighboring frame to generate the synthesized image $\mathcal{\Tilde{I}}$ given the disparity estimation $d$ and relative camera pose $\theta$. In monocular depth networks, only $\mathcal{I}$ is used to predict disparity. In stereo depth networks, both images are used to predict disparity.}
    \label{fig:workflow}
\end{figure}

Self-supervised learning methods, which construct auxiliary objectives for training models, are often used to overcome the lack of ground truth depth information.
Most work in self-supervised learning for depth estimation synthesizes images from different, but neighboring, viewpoints into one common frame, given estimated depth and ego motion, and maximizes the similarity between observed and synthesized frames in place of ground truth supervision. The learning task is shown in Figure~\ref{fig:workflow}. Given a disparity prediction (which is proportional to inverse depth, we use the terms interchangeably) from a primary coordinate frame and relative camera pose, the image from a neighboring viewpoint is warped to the primary frame, followed by the computation of a synthesis loss based on the appearance difference between the observed and synthesized images. The ideas behind this self-supervised approach are valid given that neighboring frames only differ in viewpoint. This paradigm creates new opportunities to train depth estimation networks in previously data-constrained domains and has led to many top-performing self-supervised depth estimation networks \citep{xie2016deep3d, godard2019digging,tonioni2019real}.

However, during internal experimentation using the image synthesis loss for refining depth estimation networks after transfer to different data domains, we observed divergence (formal definition in \autoref{ssec:divergence}) between performance on the surrogate (image synthesis) and true target objective (depth estimation). We observed that even if in some cases the networks final performance improved, optimization failed to stabilize for the most optimal operating point found; in other cases, the depth estimation performance of networks directly worsens as image synthesis improves. This circumstance is worrisome because in the faithful self-supervised learning setting, \emph{where true depth measurements are unknown}, a well-intended attempt at improving depth estimation performance of a transferred model through fine-tuning using image synthesis may, in fact, result in the exact opposite: worsening of depth estimation performance. Therefore, in this paper, we raise the following questions: 
\begin{enumerate}
    \item \textit{Is image synthesis effective as an auxiliary task for training deep depth estimation networks?} Yes, but only to a certain extent. We conduct extensive experiments using various contemporary networks on several datasets. We show in \autoref{ssec:test} that the gradient from image synthesis will actually \textit{disagree} with that of the theoretically \textit{unobserved} depth loss after a certain point. Thus, updating parameters in the negative gradient direction determined from the image synthesis loss will instead worsen depth prediction. The divergence is observed both at image-level and pixel-level (\autoref{ssec:local_change}).
    
    \item \textit{Can perfect image synthesis be achieved from ground truth depth information?} In short, no. In \autoref{ssec:gt}, we demonstrate that even using the ground truth depth, image synthesis is far from perfect. This suggests that optimizing for image synthesis may indeed deviate from the true objective of depth estimation.
    
    \item \textit{What is behind the diverging phenomenon between image synthesis and depth estimation?} In \autoref{ssec:aleatoric_uncertainty}, we analyze the loss manifold of image synthesis at places where we observe divergence. We show the disagreement between image synthesis loss and depth error, and the non-convexity of image synthesis loss at these places. We attribute the source of uncertainties to the data itself, also known as aleatoric uncertainty.
    
    \item \textit{What is the larger significance of this finding?} In Appendix I, we summarize a literature review following PRISMA guidelines~\citep{moher2009preferred} that resulted in 127 papers using image synthesis loss, or small variations thereof. None of these papers reported a diverging behaviour, which underscores the impact of the image synthesis paradigm on self-supervised depth estimation. We hope that identifying this issue will enable further research on improving on the clearly impactful image synthesis paradigm for self-supervised training.
\end{enumerate}

\section{Background and Related Work}
\subsection{Learning-based Depth Estimation}
Recently, deep learning has shown promising performance for depth estimation, where depth can be estimated from monocular image or stereo image pairs. In both cases, the depth estimation $Z$ is calculated from the disparity estimation $d$, with $Z \propto 1/d$. 
Monocular depth networks rely on data priors to regress up-to-scale depth from a single image~\citep{dijk2019neural}, \textit{i.e.} the exact proportionality between depth and disparity are unknown; while stereo depth networks rely on pixel matching between image pairs to produce metric depth estimation~\citep{szeliski2010computer}, \textit{i.e.} depth is found exactly with $Z=bf/d$, where $b$ is the stereo baseline and $f$ is the focal length. Both monocular and stereo networks can be trained in a self-supervised way using image synthesis (\autoref{fig:workflow}).

To ensure that our findings regarding the image synthesis loss are not limited to a specific design of the network, we have selected a diverse set of architectures, including both monocular networks and stereo networks. Due to the high sensitivity to camera setup and scene variation for monocular depth networks \citep{dijk2019neural}, we only included the SOTA MonoDepth2~\citep{godard2019digging}. As stereo networks are better constrained in terms of depth estimation, we included several recent networks of different network design, including convolution networks MADNet~\citep{tonioni2019real}, HSM~\citep{yang2019hierarchical}, GwcNet~\citep{guo2019group}, and a transformer network STTR~\citep{li2021revisiting}. We refer interested readers to Appendix A and the papers for more details.

\subsection{Image Synthesis}
As illustrated in \autoref{fig:workflow}, a synthesized image $\mathcal{\Tilde{I}}$ can be generated given the disparity, relative camera pose, and the image from a neighboring viewpoint $\mathcal{I}_N$. This can be achieved by either using images from concurrent observations from different cameras (\textit{spatial stereo}), or a series of observations from the same camera (\textit{temporal stereo}). In both cases, image synthesis is only valid for co-observed and static objects \citep{godard2019digging}. In our experiment, we only use the valid pixels and the ground truth camera poses to constrain the varying factor of image synthesis to be only the estimated disparity. To enable gradients to flow backwards through the disparity prediction to the network parameters, differentiable warping~\citep{jaderberg2015spatial} is used for synthesizing images.

\subsection{Self-supervised Training Using Image Synthesis for Depth Estimation} 
\label{ssec:syn_loss}
The most commonly used image synthesis loss was first proposed for image denoising and demosaicing tasks by Zhao \textit{et al.}~\citep{zhao2015loss} and later adapted for depth estimation in Godard \textit{et al}~\citep{godard2017unsupervised}. The loss itself is typically formulated as a weighted sum of structural similarity (SSIM) 
loss~\citep{wang2004image} and L1 loss between the observed and synthesized images, which can be written as 
\begin{align}
\label{eq:synthloss}
    L_{syn} &= \alpha L_{SSIM} + (1-\alpha) L_{L1} \,, \\ \label{eqn:ssim_loss}
    L_{SSIM} &= \frac{1}{2|\mathcal{N}|}\sum_{i\in \mathcal{N}}1 - \frac{2\mu(\mathcal{I}_i)\mu(\mathcal{\Tilde{I}}_i)+C_1}{\mu^2(\mathcal{I}_i)+\mu^2(\mathcal{\Tilde{I}}_i)+C_1} \cdot \frac{2\sigma(\mathcal{I}_i)\sigma(\mathcal{\Tilde{I}}_i)+C_2}{\sigma^2(\mathcal{I}_i)+\sigma^2(\mathcal{\Tilde{I}}_i)+C_2} \,, \\ 
    L_{L1} &= \frac{1}{|\mathcal{N}|}\sum_{i\in \mathcal{N}}|\mathcal{I}_i - \mathcal{\Tilde{I}}_i| \,.
\end{align}

\noindent $\mathcal{I}_i,\mathcal{\Tilde{I}}_i$ are the $i$-th pixels in the total $\mathcal{N}$ pixels. In \autoref{eqn:ssim_loss}, $\mu$ and $\sigma$ are the mean and standard deviation within a local patch, $C_1$ and $C_2$ are small constants to avoid division by 0. The synthesis loss is designed to measure both shape and appearance similarities. SSIM is mean balanced; therefore, it better preserves high frequency signals and is robust to chromatic differences. 
Conversely, L1 preserves color and luminance regardless of local structure. $\alpha \in [0,1]$ is a weighting factor measuring the trade-off between the two loss terms. 

\subsection{Additional Regularization and Auxiliaries}
It has previously been observed that $L_{syn}$ suffers from a gradient locality problem, where both depth loss and \syn~stagnate, since it relies on the comparison of localized pixel intensities~\citep{sharma2019unsupervised,yin2018geonet}. Multi-scale regularization, \textit{i.e.}, estimating disparities and computing $L_{syn}$ at different spatial scales ($L_{ms}$), has been reported to mitigate the gradient locality problem~\citep{sharma2019unsupervised,godard2017unsupervised,godard2019digging}. Additionally, an edge-aware smoothness regularizing term $L_{smth}$ on the disparity map has also been included~\citep{li2018occlusion,tonioni2019real} to encourage smoothness of the estimation,
\begin{equation}
    L_{smth} = \frac{1}{|\mathcal{N}|} \sum_{i\in \mathcal{N}} |\delta_x d_i|e^{-|\delta_x \mathcal{I}_i|} + |\delta_y d_i|e^{-|\delta_y \mathcal{I}_i|} \,.
\end{equation}
Here $d$ is the estimated disparity map, and $\delta$ denotes first order gradient. Due to the popularity of these two regularization approaches, we conduct additional experiments and show in Appendix B neither of these regularization approaches prevents the divergent phenomenon observed (often only mitigating it to a negligible extent).

Other auxiliary techniques to $L_{syn}$ have also been proposed. For example, \citep{wang2020faster} uses deep reinforcement learning to decide if the gradient computed from $L_{syn}$ for a specific stereo pair of images should back-propagate to the network. In~\citep{zhong2018open}, a recurrent neural network is used to backpropagate the gradient of $L_{syn}$ conditioned on a sequence of past frames. We do not consider these works in our experiments because they add additional sophistication and challenges associated that would be conflated with the divergence issue under examination.

\subsection{Divergence and Spearman's Rank Coefficients}
\label{ssec:divergence}
\begin{figure}
    \centering
    \subfloat[$r_s=+1$]{\includegraphics[height=0.22\linewidth,trim={1cm 0cm 2cm 0cm},clip]{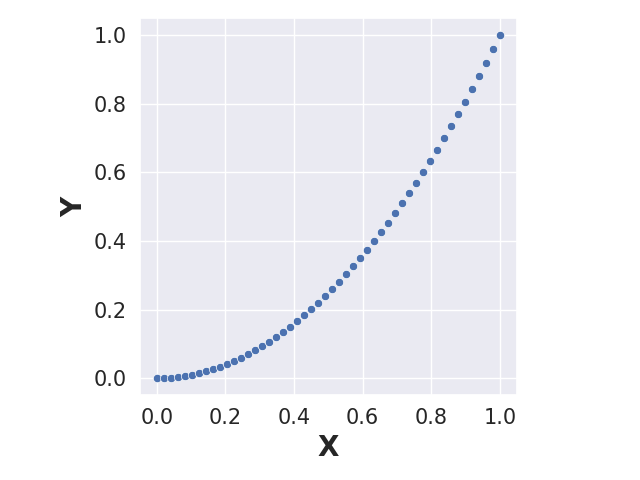}}
    \subfloat[$r_s=0.00$]{\includegraphics[height=0.22\linewidth,trim={1cm 0cm 2cm
    0cm},clip]{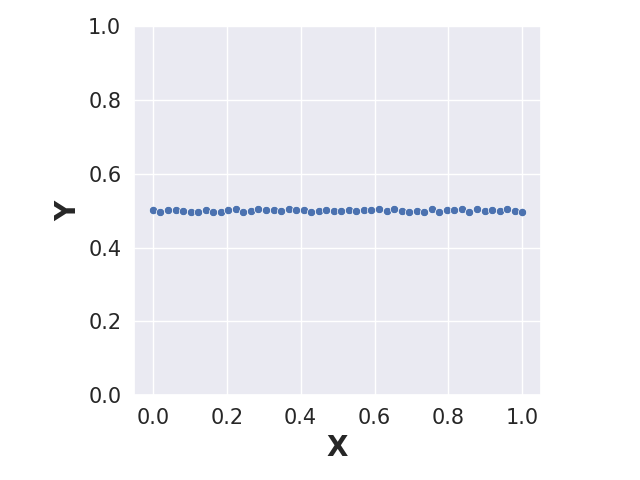}}
    \subfloat[$r_s=0.00$]{\includegraphics[height=0.22\linewidth,trim={1cm 0cm 2cm
    0cm},clip]{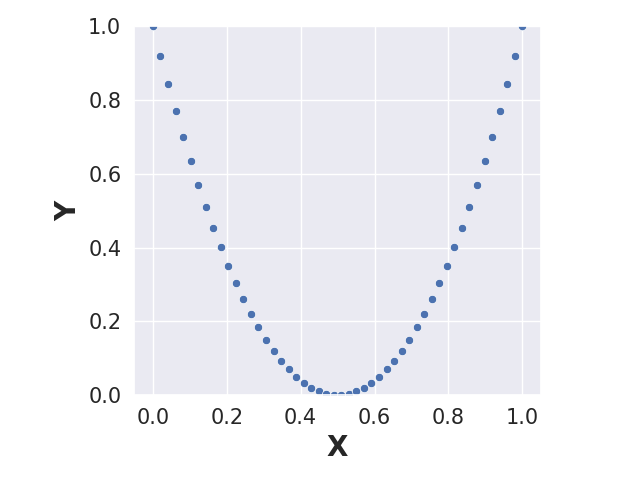}}
    \subfloat[$r_s=-1$]{\includegraphics[height=0.22\linewidth,trim={1cm 0cm 2cm 0cm},clip]{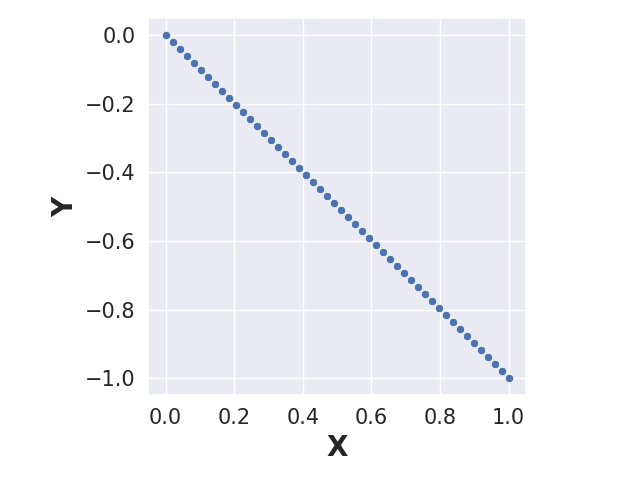}}
    
    \caption{Examples of Spearman's rank coefficients between two variables $X, Y$. (a) Positive, (b-c) weak and (d) negative monotonic relationship.}
    \label{fig:spearman_toy}
\end{figure}

In our experiment, we report the accuracy of disparity estimation via end-point-error (EPE), \textit{i.e.}, the absolute disparity error. Since the EPE and \syn~are of different units, we quantitatively assess the divergence between \syn~and EPE as their \textit{monotonic} relationship, which is quantified using the Spearman's Rank Coefficients $r_s$ \citep{ramsey1989critical} due to its robustness against outliers and noise \citep{schober2018correlation}. Given two sets of variables of size $N$ that are co-observed, $X=\{x_i : i \in[1,N] \}, Y=\{y_i : i \in [1,N] \}$, 
\begin{equation}
    r_s = \frac{\frac{1}{N}\sum_{i=1}^N(R_{x_i} - \bar{R}_x)\cdot(R_{y_i} - \bar{R}_y)}{\sqrt{(\frac{1}{N}\sum_{i=1}^N R_{x_i} - \bar{R}_x) \cdot (\frac{1}{N}\sum_{i=1}^N R_{y_i} - \bar{R}_y)}}
    \label{eqn:spearman}
\end{equation}
where $R$ is the rank and $\bar{R}$ is the mean rank. In \autoref{eqn:spearman}, the numerator is the cross-covariance between $R_X,R_Y$ while the denominator is the product of the covariances of $R_X,R_Y$. The coefficient $r_s$ ranges from -1 to +1, with -1 indicating exact negative monotonic trend, +1 indicating exact positive monotonic trend, and 0 indicating no correlation as shown in \autoref{fig:spearman_toy}. Significance of $r_s$ is reported via critical probability (p) values. We introduce the following definition of \textit{divergence}:
\begin{definition}
\label{def:divergence}
Given two sets of variables of size $N$, $X=\{x_i : i \in[1,N] \}, Y=\{y_i : i \in [1,N]\}$, a divergence is observed between $X, Y$ if both conditions are met
\begin{enumerate}
    \item $r_s < 0.39$: where a weak or negative correlation between the two variables is observed, 
    \item $p < 0.05$: the result is statistically significant.
\end{enumerate}
\end{definition}
The above thresholds are adopted from the commonly established values \citep{guilford1950fundamental,rowntree1981statistics,de2014basic}. We show in Appendix C that other thresholds used in different specialities do not affect the result significantly. Intuitively, divergence occurs when \syn~decreases while EPE 1) stagnates shown in \autoref{fig:spearman_toy} (b), 2) decreases then increases (or vice versa) shown in \autoref{fig:spearman_toy} (b), or 3) increases shown in \autoref{fig:spearman_toy} (d).

\subsection{Aleatoric Uncertainty}
Uncertainty is an active topic in machine/deep learning. In general, uncertainty can be modeled as \textit{epistemic} and \textit{aleatoric} uncertainties \citep{der2009aleatory}. Epistemic uncertainty refers to the ignorance of the model, for example, the limited capacity of a specific network architecture design. On the other hand, aleatoric uncertainty refers to the randomness of the problem or data itself. An example of aleatoric uncertainty is the coin flipping problem \citep{hullermeier2021aleatoric}, where even if the best model is given, we still cannot obtain a definite head/tail prediction of next coin flip. We refer interested readers to \citep{hullermeier2021aleatoric} for in-depth review of relevant literature. In this work, we try our best to reduce epistemic uncertainty by using various contemporary networks and ensure enough training of each network onto the target dataset, which leaves the observed divergence between image synthesis and depth estimation to aleatoric uncertainties originated from the training data.

In depth estimation literature, prior work such as \citep{godard2019digging,luo2019every,vijayanarasimhan2017sfm} model aleatoric uncertainties with a pixel-wise mask to exclude certain pixels from the training process for better performance. The excluded pixels include occluded pixels that are not commonly observed by two frames and pixels of dynamic objects that break the static scene assumption. However, none of the prior work reported the diverging phenomenon even \textit{after excluding} these pixels.

\section{Experimental Setup}
Instead of only reporting the final validation performance like prior work, we monitor how \syn~and EPE change respectively at each epoch to check if there is a divergence between \syn~and EPE.

\subsection{Dataset} 
We conduct experiments on four datasets to demonstrate that the disagreement between $L_{syn}$ and depth prediction error is not limited to a single domain. The datasets of interest provide varied image statistics and scene complexity. Furthermore, the ground truth data are captured with different depth modalities, which constitutes a stable backdrop for observing the effects of self-supervision. The datasets selected have corresponding ground truth depth, which can be used to monitor divergence. Since real datasets may have chromatic differences between left/right images, we align the left and right images appearance using a per-channel shift \citep{jin2001real}. We normalize the images to 0-1 when computing \syn~to make SSIM valid \citep{wang2004image}. More details about dataset used can be found in Appendix A.

\subsubsection{Spatial Stereo Dataset}
KITTI2015~\citep{menze2015object} contains 200 pairs of images of street scenes of resolution $1242\times375$ with \textit{sparse} ground truth disparity maps computed from LiDAR. We use the first 150 images for training and the rest 50 images for validation. SERV-CT~\citep{eddie2020serv} contains 16 pairs of \textit{ex vivo} endoscopic images of resolution $720\times576$ with \textit{dense} ground truth disparity maps computed from an aligned CT scan. We use the first 8 images for training and the remaining 8 images for validation. Middlebury2014 Q \citep{scharstein2014high} contains 15 images of indoor scenes of various image resoltuion with \textit{dense} disparity maps acquired via structured light. We use the first 10 images for training and the remaining 5 images for validation. 

\subsubsection{Temporal Stereo Dataset}
Since dynamic objects violate the image synthesis assumption, we use the synthetic New Tsukuba Stereo Dataset fluorescent \citep{martull2012realistic} which is static indoor scenes with associated ground truth camera pose and \textit{dense} disparity maps. The resolution is $640\times480$. We use the first 130 frames for training and the remaining 50 images for validation. Using this dataset allows us to investigate the temporal effect of self-supervised depth estimation.

\subsection{Hyperparameters} 
We use the pre-trained weights provided by the authors (see Appendix A for more details). We scale the originally used learning rate by 0.1 as the pre-trained weights are well-initialized. We use the AdamW optimizer with a weight decay of 1e-4~\citep{loshchilov2017decoupled}. We demonstrate the choice of optimizer does not fundamentally change the result (see Appendix G).

We follow prior work~\citep{godard2017unsupervised,tonioni2019real} and set $\alpha=0.85$ in \autoref{eq:synthloss}, and set $C_1=1$e-4, $C_2=9$e-4, the kernel size used to compute $\mu,\sigma$ to $3\times3$ in \autoref{eqn:ssim_loss}. We also show that these hyper-parameters ($\alpha$ and kernel size) marginally affect the result but do not resolve the divergence issue (see Appendix E and F). Following~\citep{godard2019digging}, we upsample disparities from each scale to full resolution and set weightings for \syn~from each scale to 1, and weighting for \lsmth~to 1e-2. We train for 200 epochs on all dataset. We run all experiments on one Nvidia Titan RTX GPU.

\section{Results and Discussion}
\subsection{Divergence in validation}
\label{ssec:test}
The quantitative result is summarized in \autoref{tab:validation}. Comparing the change of EPE from final to initial values, some networks are able to improve on certain dataset (highlighted as {\color[HTML]{002060} \textbf{blue}}), which agrees with the conclusion of prior work that image synthesis improves the depth estimation. However, many networks have worse performance after optimizing for image-synthesis (highlighted as {\color[HTML]{C00000} \textbf{red}}). Furthermore, we note that the $r_s$ is not always a positive number and often falls below the weak correlation threshold (highlighted as {\color[HTML]{C00000} \textbf{red}}), which indicates that EPE does not strictly follow the \syn~monotonically. The only strongly correlated case without any divergence is HSM trained on Middlebury2014. Fraction of divergence is also reported, where on average 0.54 of the data across different networks and dataset exhibit divergence. 

\begin{figure}[htpb]
    \centering
    \includegraphics[width=\textwidth]{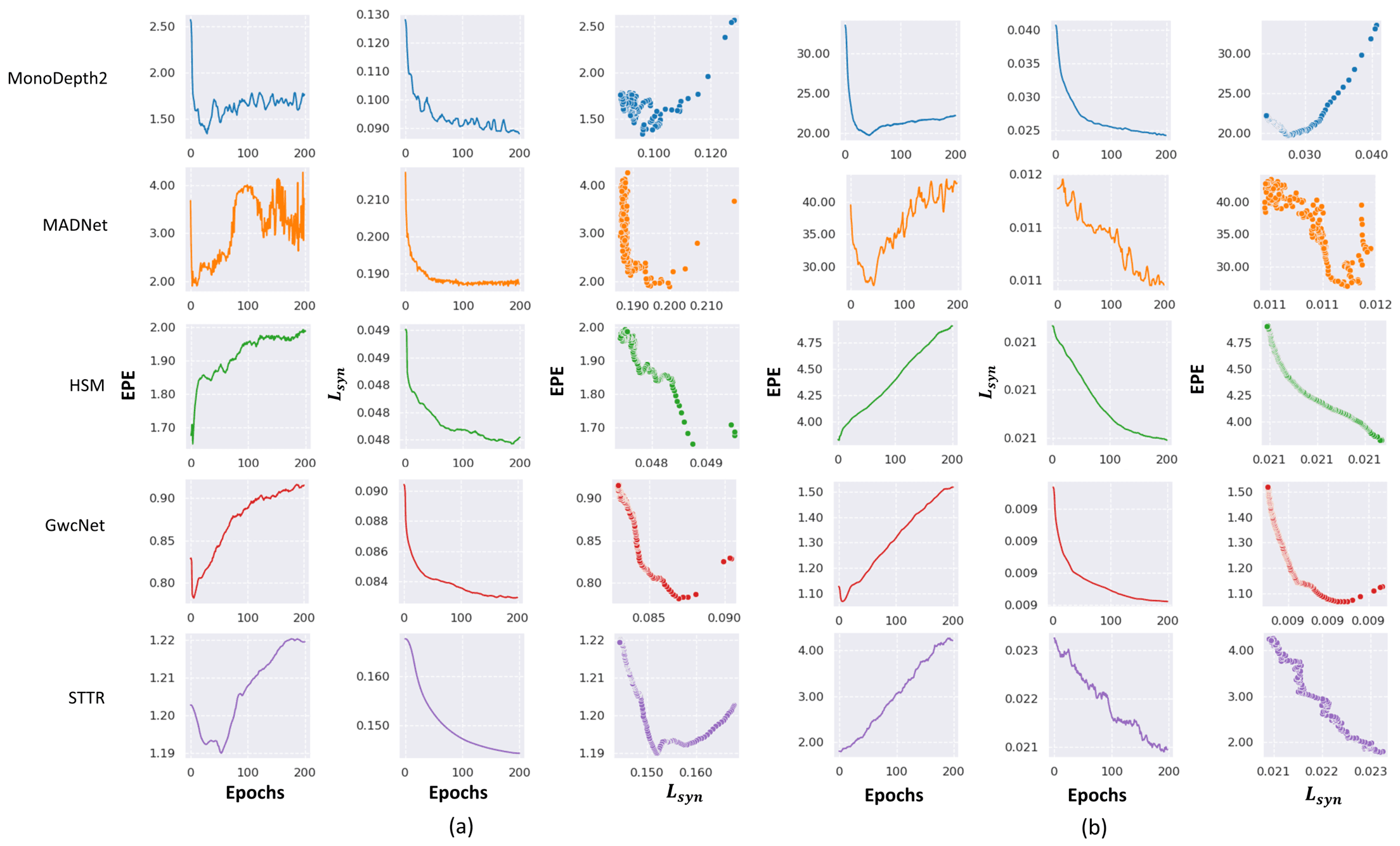}
    \caption{Validation result as training proceeds of (a) KITTI2015 dataset and (b) SERV-CT dataset.}
    \label{fig:validation}
\end{figure}

We plot examples of divergence from two datasets in \autoref{fig:validation}. It is common to many networks that there exist a positive correlation between EPE and \syn~when \syn~is large, and after reaching the minima, the network worsens w.r.t depth afterwards. In other cases, the best operating point found is the starting point. This observation suggests a practical upper bound on depth estimation performance when optimizing for image synthesis: as a network improves on the target domain, further optimization of image synthesis can be harmful where networks fail to stabilize at the best optimal operating point found. Lastly, divergence is found in both spatial and temporal stereo dataset. We also observe divergence of \syn~and EPE when evaluated on the training data. Details can be found in Appendix B.

\begin{table}[htpb]
\centering
\caption{Validation result on four datasets. $\Delta$: changes between final and initial EPE ({\color[HTML]{002060} \textbf{blue}} indicates EPE decreases; {\color[HTML]{C00000} \textbf{red}} indicates EPE increases). $r_s$: Spearman's Rank Coefficient ({\color[HTML]{002060} \textbf{blue}} indicates average above 0.39; {\color[HTML]{C00000} \textbf{red}} indicates average below 0.39). Divergence: fraction of diverging training instances observed based on \autoref{def:divergence}.}
\label{tab:validation}
\resizebox{0.85\textwidth}{!}{%
\begin{tabular}{ccccccc}
\hline
\multicolumn{1}{|c|}{} & \multicolumn{1}{c|}{} & \multicolumn{3}{c|}{\textbf{EPE}} & \multicolumn{1}{c|}{} & \multicolumn{1}{c|}{} \\ \cline{3-5}
\multicolumn{1}{|c|}{\multirow{-2}{*}{\textbf{Dataset}}} & \multicolumn{1}{c|}{\multirow{-2}{*}{\textbf{Network}}} & \multicolumn{1}{c|}{\textbf{Initial}} & \multicolumn{1}{c|}{\textbf{Final}} & \multicolumn{1}{c|}{\textbf{$\Delta$}} & \multicolumn{1}{c|}{\multirow{-2}{*}{\textbf{$r_s$}}} & \multicolumn{1}{c|}{\multirow{-2}{*}{\textbf{Divergence}}} \\ \hline
\multicolumn{1}{|c|}{} & \multicolumn{1}{c|}{MonoDepth2} & \multicolumn{1}{c|}{6.24$\pm$ 3.49} & \multicolumn{1}{c|}{4.07$\pm$ 2.72} & \multicolumn{1}{c|}{{\color[HTML]{002060} -2.18 (-35\%)}} & \multicolumn{1}{c|}{{\color[HTML]{002060} +0.41$\pm$ 0.38}} & \multicolumn{1}{c|}{0.20} \\ \cline{2-7} 
\multicolumn{1}{|c|}{} & \multicolumn{1}{c|}{MADNet} & \multicolumn{1}{c|}{8.28$\pm$ 5.83} & \multicolumn{1}{c|}{1.92$\pm$ 1.70} & \multicolumn{1}{c|}{{\color[HTML]{002060} -6.35 (-73\%)}} & \multicolumn{1}{c|}{{\color[HTML]{C00000} +0.35$\pm$ 0.45}} & \multicolumn{1}{c|}{0.36} \\ \cline{2-7} 
\multicolumn{1}{|c|}{} & \multicolumn{1}{c|}{HSM} & \multicolumn{1}{c|}{1.31$\pm$ 0.42} & \multicolumn{1}{c|}{1.27$\pm$ 0.40} & \multicolumn{1}{c|}{{\color[HTML]{002060} -0.04 (-3\%)}} & \multicolumn{1}{c|}{{\color[HTML]{C00000} +0.04$\pm$ 0.60}} & \multicolumn{1}{c|}{0.48} \\ \cline{2-7} 
\multicolumn{1}{|c|}{} & \multicolumn{1}{c|}{GwcNet} & \multicolumn{1}{c|}{1.27$\pm$ 0.44} & \multicolumn{1}{c|}{1.37$\pm$ 0.47} & \multicolumn{1}{c|}{{\color[HTML]{C00000} +0.10 (+9\%)}} & \multicolumn{1}{c|}{{\color[HTML]{C00000} -0.41$\pm$ 0.58}} & \multicolumn{1}{c|}{0.82} \\ \cline{2-7} 
\multicolumn{1}{|c|}{\multirow{-5}{*}{KITTI2015}} & \multicolumn{1}{c|}{STTR} & \multicolumn{1}{c|}{1.50$\pm$ 0.81} & \multicolumn{1}{c|}{2.53$\pm$ 2.79} & \multicolumn{1}{c|}{{\color[HTML]{C00000} +1.03 (+66\%)}} & \multicolumn{1}{c|}{{\color[HTML]{C00000} -0.01$\pm$ 0.21}} & \multicolumn{1}{c|}{0.40} \\ \hline
\multicolumn{1}{l}{} &  &  &  &  &  &  \\ \hline
\multicolumn{1}{|c|}{} & \multicolumn{1}{c|}{MonoDepth2} & \multicolumn{1}{c|}{28.18$\pm$ 5.31} & \multicolumn{1}{c|}{11.85$\pm$ 5.39} & \multicolumn{1}{c|}{{\color[HTML]{002060} -16.33 (-59\%)}} & \multicolumn{1}{c|}{{\color[HTML]{002060} +0.65$\pm$ 0.51}} & \multicolumn{1}{c|}{0.25} \\ \cline{2-7} 
\multicolumn{1}{|c|}{} & \multicolumn{1}{c|}{MADNet} & \multicolumn{1}{c|}{57.71$\pm$ 23.67} & \multicolumn{1}{c|}{35.10$\pm$ 6.23} & \multicolumn{1}{c|}{{\color[HTML]{002060} -22.61 (-18\%)}} & \multicolumn{1}{c|}{{\color[HTML]{C00000} +0.34$\pm$ 0.60}} & \multicolumn{1}{c|}{0.50} \\ \cline{2-7} 
\multicolumn{1}{|c|}{} & \multicolumn{1}{c|}{HSM} & \multicolumn{1}{c|}{2.38$\pm$ 0.72} & \multicolumn{1}{c|}{2.79$\pm$ 1.00} & \multicolumn{1}{c|}{{\color[HTML]{C00000} +0.42 (+18\%)}} & \multicolumn{1}{c|}{{\color[HTML]{C00000} -0.29$\pm$ 0.67}} & \multicolumn{1}{c|}{0.62} \\ \cline{2-7} 
\multicolumn{1}{|c|}{} & \multicolumn{1}{c|}{GwcNet} & \multicolumn{1}{c|}{3.53$\pm$ 1.89} & \multicolumn{1}{c|}{2.73$\pm$ 0.82} & \multicolumn{1}{c|}{{\color[HTML]{002060} -0.80 (-7\%)}} & \multicolumn{1}{c|}{{\color[HTML]{C00000} -0.58$\pm$ 0.56}} & \multicolumn{1}{c|}{0.88} \\ \cline{2-7} 
\multicolumn{1}{|c|}{\multirow{-5}{*}{SERV-CT}} & \multicolumn{1}{c|}{STTR} & \multicolumn{1}{c|}{3.70$\pm$ 2.06} & \multicolumn{1}{c|}{4.56$\pm$ 3.05} & \multicolumn{1}{c|}{{\color[HTML]{C00000} +0.86 (+20\%)}} & \multicolumn{1}{c|}{{\color[HTML]{C00000} -0.73$\pm$ 0.38}} & \multicolumn{1}{c|}{1.00} \\ \hline
\multicolumn{1}{l}{} & \multicolumn{1}{l}{} & \multicolumn{1}{l}{} & \multicolumn{1}{l}{} & \multicolumn{1}{l}{} &  &  \\ \hline
\multicolumn{1}{|c|}{} & \multicolumn{1}{c|}{MonoDepth2} & \multicolumn{1}{c|}{17.07$\pm$ 8.52} & \multicolumn{1}{c|}{12.83$\pm$ 8.58} & \multicolumn{1}{c|}{{\color[HTML]{002060} -4.24 (-28\%)}} & \multicolumn{1}{c|}{{\color[HTML]{C00000} -0.28$\pm$ 0.83}} & \multicolumn{1}{c|}{0.60} \\ \cline{2-7} 
\multicolumn{1}{|c|}{} & \multicolumn{1}{c|}{MADNet} & \multicolumn{1}{c|}{13.06$\pm$ 4.50} & \multicolumn{1}{c|}{14.58$\pm$ 3.88} & \multicolumn{1}{c|}{{\color[HTML]{C00000} +1.52 (+18\%)}} & \multicolumn{1}{c|}{{\color[HTML]{C00000} -0.29$\pm$ 0.51}} & \multicolumn{1}{c|}{0.60} \\ \cline{2-7} 
\multicolumn{1}{|c|}{} & \multicolumn{1}{c|}{HSM} & \multicolumn{1}{c|}{2.60$\pm$ 1.57} & \multicolumn{1}{c|}{2.10$\pm$ 1.40} & \multicolumn{1}{c|}{{\color[HTML]{002060} -0.49 (-21\%)}} & \multicolumn{1}{c|}{{\color[HTML]{002060} +0.89$\pm$ 0.17}} & \multicolumn{1}{c|}{0.00} \\ \cline{2-7} 
\multicolumn{1}{|c|}{} & \multicolumn{1}{c|}{GwcNet} & \multicolumn{1}{c|}{3.07$\pm$ 4.00} & \multicolumn{1}{c|}{3.21$\pm$ 4.17} & \multicolumn{1}{c|}{{\color[HTML]{C00000} +0.14 (+4\%)}} & \multicolumn{1}{c|}{{\color[HTML]{002060} +0.42$\pm$ 0.60}} & \multicolumn{1}{c|}{0.20} \\ \cline{2-7} 
\multicolumn{1}{|c|}{\multirow{-5}{*}{Middlebury2014}} & \multicolumn{1}{c|}{STTR} & \multicolumn{1}{c|}{1.69$\pm$ 2.66} & \multicolumn{1}{c|}{1.33$\pm$ 1.24} & \multicolumn{1}{c|}{{\color[HTML]{002060} -0.35 (-5\%)}} & \multicolumn{1}{c|}{{\color[HTML]{C00000} -0.58$\pm$ 0.65}} & \multicolumn{1}{c|}{0.87} \\ \hline
\multicolumn{1}{l}{} & \textbf{} & \textbf{} & \textbf{} & \textbf{} &  &  \\ \hline
\multicolumn{1}{|c|}{} & \multicolumn{1}{c|}{MonoDepth2} & \multicolumn{1}{c|}{10.74$\pm$ 4.09} & \multicolumn{1}{c|}{8.45$\pm$ 3.21} & \multicolumn{1}{c|}{{\color[HTML]{002060} -2.29 (-20\%)}} & \multicolumn{1}{c|}{{\color[HTML]{C00000} +0.26$\pm$ 0.38}} & \multicolumn{1}{c|}{0.38} \\ \cline{2-7} 
\multicolumn{1}{|c|}{} & \multicolumn{1}{c|}{MADNet} & \multicolumn{1}{c|}{26.13$\pm$ 11.99} & \multicolumn{1}{c|}{16.57$\pm$ 6.20} & \multicolumn{1}{c|}{{\color[HTML]{002060} -9.56 (-30\%)}} & \multicolumn{1}{c|}{{\color[HTML]{C00000} 0.00$\pm$ 0.42}} & \multicolumn{1}{c|}{0.56} \\ \cline{2-7} 
\multicolumn{1}{|c|}{} & \multicolumn{1}{c|}{HSM} & \multicolumn{1}{c|}{1.66$\pm$ 0.47} & \multicolumn{1}{c|}{1.62$\pm$ 0.39} & \multicolumn{1}{c|}{{\color[HTML]{002060} -0.04 (-1\%)}} & \multicolumn{1}{c|}{{\color[HTML]{C00000} +0.15$\pm$ 0.47}} & \multicolumn{1}{c|}{0.48} \\ \cline{2-7} 
\multicolumn{1}{|c|}{} & \multicolumn{1}{c|}{GwcNet} & \multicolumn{1}{c|}{1.69$\pm$ 1.14} & \multicolumn{1}{c|}{1.79$\pm$ 1.06} & \multicolumn{1}{c|}{{\color[HTML]{C00000} +0.09 (+11\%)}} & \multicolumn{1}{c|}{{\color[HTML]{C00000} -0.68$\pm$ 0.21}} & \multicolumn{1}{c|}{0.56} \\ \cline{2-7} 
\multicolumn{1}{|c|}{\multirow{-5}{*}{Tsukuba}} & \multicolumn{1}{c|}{STTR} & \multicolumn{1}{c|}{2.01$\pm$ 2.67} & \multicolumn{1}{c|}{21.95$\pm$ 7.98} & \multicolumn{1}{c|}{{\color[HTML]{C00000} +19.94 (+1458\%)}} & \multicolumn{1}{c|}{{\color[HTML]{C00000} -0.89$\pm$ 0.16}} & \multicolumn{1}{c|}{1.00} \\ \hline
 &  &  &  & \multicolumn{1}{c|}{} & \multicolumn{1}{c|}{\textbf{Average}} & \multicolumn{1}{c|}{0.54} \\ \cline{6-7} 
\end{tabular}%
}
\end{table}

\subsection{Change of EPE with decreasing \syn}
\label{ssec:local_change}
Instead of comparing the aggregated \syn~and EPE for the whole image, we instead compute the aggregated change of EPE only on pixels where \syn~is decreasing. This further isolates the inconsistent behavior. The quantitative result is summarized in \autoref{tab:local-comparison-epe-syn}. We note that even though in most cases the average $\Delta$EPE is negative (\textit{i.e.} a smaller error after training), the standard deviation is much larger, indicating many pixels have increasing error. We further highlight in red five cases where on average these pixels have increasing EPE. Therefore, minimization of \syn~is indeed not equivalent to minimization of EPE despite the theoretical justification of the image synthesis paradigm. Qualitative results can be found in Appendix D.

\begin{table}[htpb]
\centering
\caption{Changes of EPE of pixels where \syn~decreases.}
\label{tab:local-comparison-epe-syn}
\resizebox{0.64\textwidth}{!}{%
\begin{tabular}{|c|c|c|c|c|}
\hline
 & \multicolumn{4}{c|}{\textbf{$\Delta$ EPE}} \\ \cline{2-5} 
\multirow{-2}{*}{\textbf{Network}} & \textbf{KITTI2015} & \textbf{SERV-CT} & \textbf{Middlebury2014} & \textbf{New Tsukuba} \\ \hline
MonoDepth2 & -1.54$\pm$ 3.98 & -24.20 $\pm$ 40.26 & -4.04$\pm$ 11.77 & -0.54$\pm$ 5.36 \\ \hline
MADNet & -1.95 $\pm$ 13.58 & -19.03 $\pm$ 27.36 & -0.60 $\pm$ 4.86 & {\color[HTML]{C00000} +0.53 $\pm$ 3.35} \\ \hline
HSM & {\color[HTML]{C00000} +1.30$\pm$ 1.67} & {\color[HTML]{C00000} +1.28 $\pm$ 4.97} & -0.66$\pm$ 1.18 & -0.70$\pm$ 3.88 \\ \hline
GwcNet & {\color[HTML]{C00000} +0.25$\pm$ 1.49} & -5.42 $\pm$ 19.26 & -0.12$\pm$ 1.66 & -0.13$\pm$ 1.75 \\ \hline
STTR & -1.06$\pm$ 3.51 & -6.18 $\pm$ 20.14 & -0.13$\pm$ 1.07 & {\color[HTML]{C00000} +1.23$\pm$ 4.11} \\ \hline
\end{tabular}%
}
\end{table}

\subsection{Ground Truth Disparity and Non-zero $L_{syn}$}
\label{ssec:gt}
\begin{figure}[htpb]
    \centering
    \includegraphics[width=0.85\textwidth]{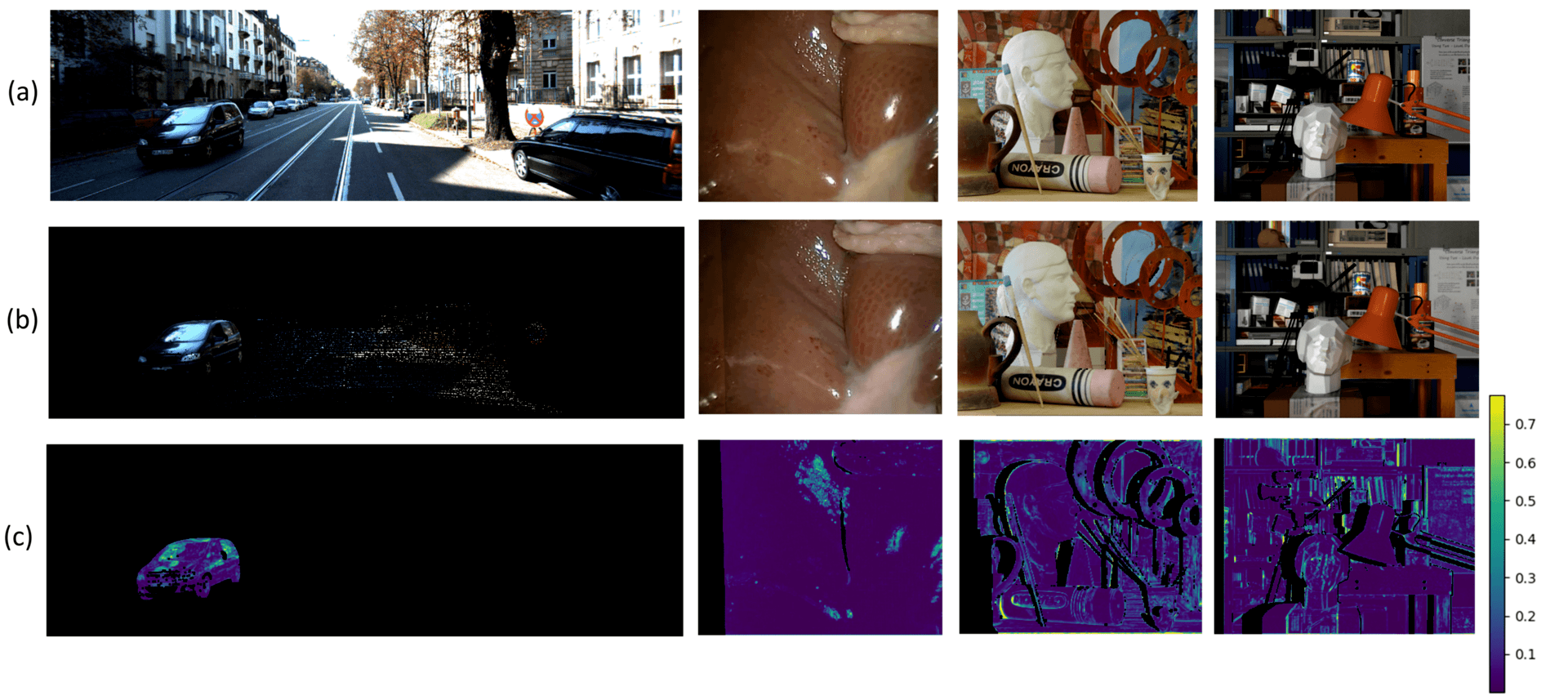}
    \caption{Visualization of (a) the input image, (b) the synthesized image, and (c) $L_{syn}$ (occluded regions or regions without valid ground truth depth data are shown in black). Left to right: KITTI2015, SERV-CT, Middlebury2014 and New Tsukuba dataset.}
    \label{fig:gt_syn_loss}
\end{figure}

In fact, we show that even with ground truth disparity, $L_{syn}$ will not be 0 (\autoref{tab:gt_syn}). Therefore, if a network hypothetically predicts perfect disparity, non-zero gradients computed for \syn~w.r.t disparity will pass to the network. In this case, increasing \syn~may cause a larger perturbation of the model weights, leading to potentially divergent behavior. Though this non-zero gradient may be averaged out during the training process, we showed in \autoref{ssec:test} that divergence occurs after all. We visualize examples in \autoref{fig:gt_syn_loss}. Despite visual similarities between the actual and synthesized images, there are non-zero errors across the whole image. Appendix E provides additional experiment showing optimizing either $L_{SSIM}$ and $L_{L1}$ can lead to divergent learning. 

\begin{table}[htpb]
\centering
\caption{\syn, \ssim~and \lone~when ground truth disparity is used.}
\label{tab:gt_syn}
\resizebox{0.59\textwidth}{!}{%
\begin{tabular}{|c|c|c|c|c|}
\hline
 & \textbf{KITTI2015} & \textbf{SERV-CT} & \textbf{Middlebury2014} & \textbf{New Tsukuba} \\ \hline
\textbf{\syn} & 0.120 & 0.023 & 0.053 & 0.035 \\ \hline
\textbf{\ssim} & 0.132 & 0.023 & 0.057 & 0.039 \\ \hline
\textbf{\lone} & 0.051 & 0.022 & 0.035 & 0.011 \\ \hline
\end{tabular}%
}
\end{table}

\subsection{Aleatoric Uncertainties of Image Synthesis}
\label{ssec:aleatoric_uncertainty}
After observing the divergence, we analyze the loss landscapes \citep{cheng2021explore} at places where divergence occurs in \autoref{fig:aleatoric}. As shown, the optimization for image synthesis on a pixel level is non-convex w.r.t depth and deviates from the ground truth location (EPE$=0$). We further elaborate below:

\begin{figure}[hptb]
    \centering
    \includegraphics[width=\textwidth]{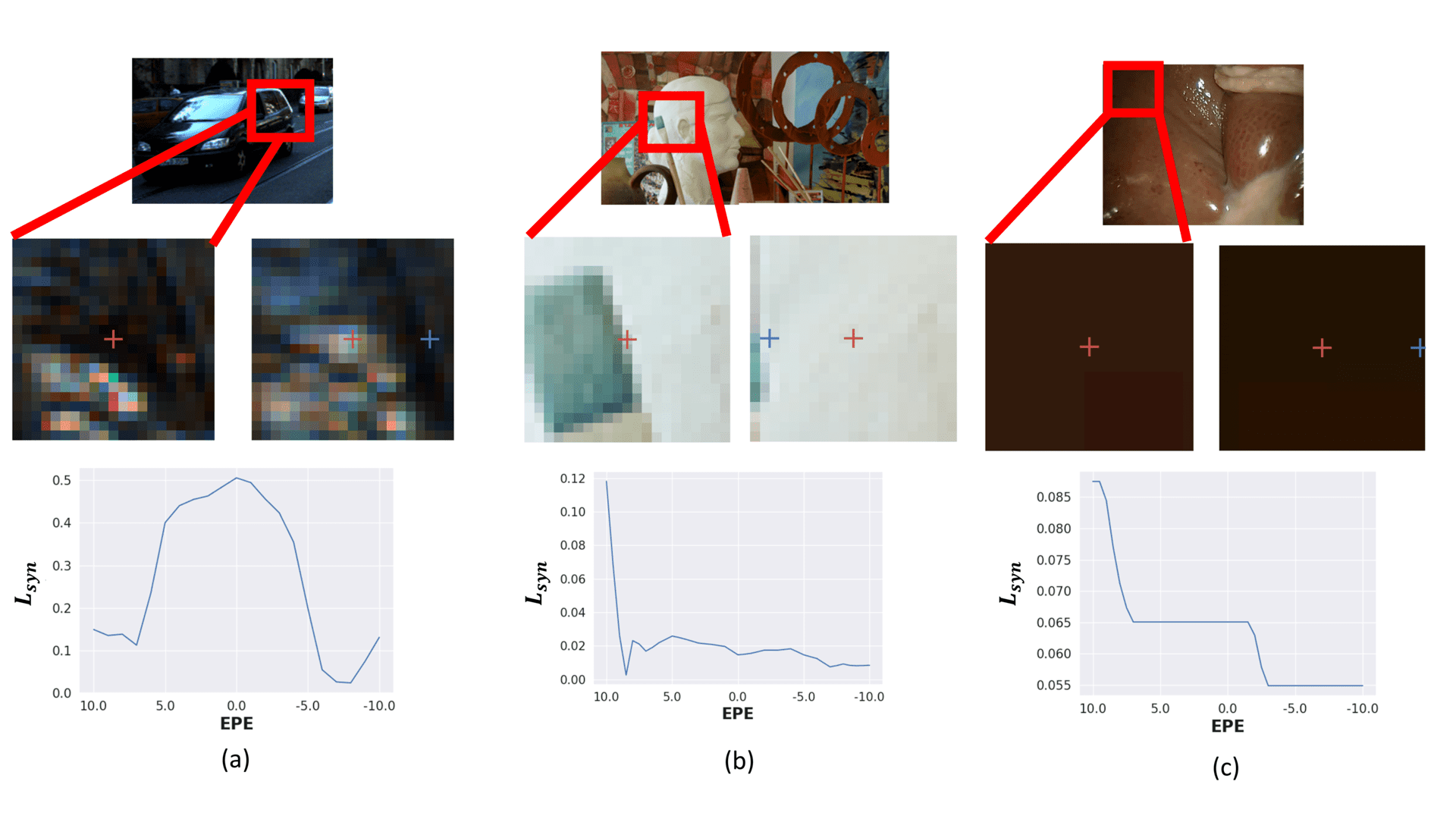}
    \caption{Aleatoric uncertainties and their associated loss landscapes. Top row: input images. Second row: zoom-in view of the primary image (left) and the neighboring image (right). Red cross indicates the corresponding pixels using ground truth depth and blue cross indicates the location of minimum \syn found within the local window. Bottom row: loss landscape of \syn~w.r.t EPE.}
    \label{fig:aleatoric}
\end{figure}

\textit{Specular reflection and shadow} - Specular reflection of non-Lambertian surfaces and shadow will occlude the true surface appearance and complicate the image synthesis process when it is observed at different positions. One example is illustrated in \autoref{fig:aleatoric} (a) where differences between the reflection in the two images will conflate with the true disparity estimation value.

\textit{Object/occlusion edge} - When a large disparity change occurs at object/occlusion edges, pixels will have different local context in primary and neighboring images, which will lead to false correspondences as shown in \autoref{fig:aleatoric} (b).

\textit{Lack of texture or repeated texture} - At a textureless (over/under exposure or uniform texture) or repeated texture area, there are many possible values that can minimize \syn, which makes image synthesis prone to error. Even if smoothness regularization is used, it does not guarantee the correctness of the disparity estimation (see video supplementary material).

\section{Significance of the Finding}
Our primary finding demonstrates that self-supervised optimization of $L_{syn}$ is limited in its ability to improve depth prediction error, especially when depth estimation has improved to a sufficient extent during self-supervised learning. While there are cases where the final performance of depth estimation on a target domain may improve (which has been the focus of most published work to date), a network may, contrary to common belief, have divergent training and fail to stabilize at the best operating point found (\autoref{fig:validation}). This finding may point to inherent limitations of contemporary self-supervised methods for depth estimation, and further, cast doubt on the use of SSIM to compare the synthesized images against the primary image as an evaluation metric for assessing depth estimation performance without ground truth \citep{gur2019single}.

\section{Conclusion and Limitations}
In this work, we have empirically demonstrated a disagreement between the image synthesis loss and the final depth estimation accuracy during self-supervised learning of depth estimation. This phenomenon provides evidence for practical limits of image synthesis for decreasing depth prediction error due to aleatoric uncertainties. We found 127 papers use this paradigm, yet none of them reported such cases. While we cannot contribute a solution to this issue at the moment, we believe that this issue warrants discussion due to the growing prevalence of image synthesis-based self-supervision. Our results currently are limited to small datasets without large scale training due to the limited availability of ground truth data. Neither do we not consider the noise and artifacts associated with the ground truth data, which may influence the result. Future work may improve upon our findings by explicitly identifying the regions which exhibit aleatoric uncertainties. Exploring additional constraints beyond image synthesis (such as geometric information \citep{wei2020deepsfm}) for training depth estimation networks and causal analysis \citep{pearl2009causality} of the aleatoric uncertainties and divergence are of interest for future work.

\bibliographystyle{plainnat}
\bibliography{reference}

\appendix
\section*{Appendix}
\section{Networks and Dataset Details}
MonoDepth2 \citep{godard2019digging} is a SOTA monocular depth estimation network. The network predicts disparity through a sigmoid layer. The pre-trained weight used is named \textit{stereo\_$640\times192$}, which is trained in a self-supervised setup using stereo images on KITTI RAW dataset \citep{geiger2013vision} (which is different from KITTI 2015 dataset). The code and pre-trained weights are released under the \textit{Monodepth v2 Liense}. 

MADNet \citep{tonioni2019real} is a stereo depth estimation network. The network has a correlation layer for predicting disparity. The pre-trained weight used is named \textit{Sythetic}, which is trained with ground truth supervision using Scene Flow \citep{mayer2016large}. The code and pre-trained weights are released under the \textit{Apache-2.0 License}.

HSM \citep{martull2012realistic} is a stereo depth estimation network. The network uses 3D convolutions for predicting disparities. The pre-trained weight used is named \textit{Middlebury model}, which is trained on customized datasets (see paper for more details). The code and pre-trained weights are released under the \textit{MIT License}.

GwcNet \citep{guo2019group} is a stereo depth estimation network. The network uses both correlation and 3D convolutions to predict disparities. The pre-trained weight used is provided in \citep{li2021revisiting} which is trained using color-augmentation on Scene Flow \citep{mayer2016large} with ground truth supervision. The code and pre-trained weights are released under the \textit{MIT License}.

STTR~\citep{li2021revisiting} is a stereo depth estimation network. The network is a transformer based network which uses attention to predict disparities. The pre-trained weight used is named \textit{STTR-light}, which is trained using color-augmentation on Scene Flow \citep{mayer2016large}. The code and pre-trained weights are released under the \textit{Apache-2.0 License}.

KITTI2015\citep{menze2015object} uses the \textit{CC BY-NC-SA 3.0 License}. SERV-CT \citep{eddie2020serv} uses the \textit{CC BY 4.0 License}. No license is provided for Middlebury2014 \citep{scharstein2014high} and New Tsukuba \citep{martull2012realistic} dataset. KITTI2015, SERV-CT and Middlebury2014 have distinct scenes for each data provided, while New Tsukuba provides a sequential observation of the same scene at 30 Hz. Due to the small camera motion for New Tsukuba, we downsample the frequency to 3 Hz.

\section{Divergence in Training}
\subsection{Divergence in Training}
\label{ssec:training}
We also monitor \syn~and EPE for training data, where no generalization is required for the network. The quantitative result is shown in \autoref{tab:training}. The disagreement of \syn~and EPE is also evident on training data as the $r_s$ rarely surpasses 0.39 and many networks have worse performance after self-supervised training. In fact, the average divergence cases of 0.61 is higher than 0.54 in the validation result (Table 1 in the manuscript), suggesting that over-optimizing for \syn~may harm the depth performance without any knowledge of the actual depth performance.

\begin{table}[htpb]
\centering
\caption{Training result on four datasets. $\Delta$: changes between final and initial EPE ({\color[HTML]{002060} \textbf{blue}} indicates EPE decreases; {\color[HTML]{C00000} \textbf{red}} indicates EPE increases). $r_s$: Spearman's Rank Coefficient ({\color[HTML]{002060} \textbf{blue}} indicates average above 0.39; {\color[HTML]{C00000} \textbf{red}} indicates average below 0.39). Divergence: fraction of diverging training instances observed based on Definition 1.}
\label{tab:training}
\resizebox{0.85\textwidth}{!}{%
\begin{tabular}{ccccccc}
\hline
\multicolumn{1}{|c|}{} & \multicolumn{1}{c|}{} & \multicolumn{3}{c|}{\textbf{EPE}} & \multicolumn{1}{c|}{} & \multicolumn{1}{c|}{} \\ \cline{3-5}
\multicolumn{1}{|c|}{\multirow{-2}{*}{\textbf{Dataset}}} & \multicolumn{1}{c|}{\multirow{-2}{*}{\textbf{Network}}} & \multicolumn{1}{c|}{\textbf{Initial}} & \multicolumn{1}{c|}{\textbf{Final}} & \multicolumn{1}{c|}{\textbf{$\Delta$}} & \multicolumn{1}{c|}{\multirow{-2}{*}{\textbf{$r_s$}}} & \multicolumn{1}{c|}{\multirow{-2}{*}{\textbf{Divergence}}} \\ \hline
\multicolumn{1}{|c|}{} & \multicolumn{1}{c|}{MonoDepth2} & \multicolumn{1}{c|}{7.20$\pm$ 3.08} & \multicolumn{1}{c|}{1.46$\pm$ 0.88} & \multicolumn{1}{c|}{{\color[HTML]{002060} -5.75 (-78\%)}} & \multicolumn{1}{c|}{{\color[HTML]{002060} +0.54$\pm$ 0.21}} & \multicolumn{1}{c|}{0.23} \\ \cline{2-7} 
\multicolumn{1}{|c|}{} & \multicolumn{1}{c|}{MADNet} & \multicolumn{1}{c|}{7.03$\pm$ 12.18} & \multicolumn{1}{c|}{1.74$\pm$ 1.64} & \multicolumn{1}{c|}{{\color[HTML]{002060} -5.29 (-65\%)}} & \multicolumn{1}{c|}{{\color[HTML]{C00000} +0.38$\pm$ 0.51}} & \multicolumn{1}{c|}{0.41} \\ \cline{2-7} 
\multicolumn{1}{|c|}{} & \multicolumn{1}{c|}{HSM} & \multicolumn{1}{c|}{1.21$\pm$ 0.49} & \multicolumn{1}{c|}{1.14$\pm$ 0.66} & \multicolumn{1}{c|}{{\color[HTML]{002060} -0.07 (-7\%)}} & \multicolumn{1}{c|}{{\color[HTML]{002060} +0.52$\pm$ 0.58}} & \multicolumn{1}{c|}{0.29} \\ \cline{2-7} 
\multicolumn{1}{|c|}{} & \multicolumn{1}{c|}{GwcNet} & \multicolumn{1}{c|}{1.31$\pm$ 2.08} & \multicolumn{1}{c|}{1.32$\pm$ 0.64} & \multicolumn{1}{c|}{{\color[HTML]{C00000} +0.01 (+4\%)}} & \multicolumn{1}{c|}{{\color[HTML]{C00000} -0.67$\pm$ 0.40}} & \multicolumn{1}{c|}{0.95} \\ \cline{2-7} 
\multicolumn{1}{|c|}{\multirow{-5}{*}{KITTI2015}} & \multicolumn{1}{c|}{STTR} & \multicolumn{1}{c|}{1.46$\pm$ 2.78} & \multicolumn{1}{c|}{2.02$\pm$ 3.29} & \multicolumn{1}{c|}{{\color[HTML]{C00000} +0.56 (+37\%)}} & \multicolumn{1}{c|}{{\color[HTML]{C00000} -0.05$\pm$ 0.24}} & \multicolumn{1}{c|}{0.99} \\ \hline
\multicolumn{1}{l}{} &  &  &  &  &  &  \\ \hline
\multicolumn{1}{|c|}{} & \multicolumn{1}{c|}{MonoDepth2} & \multicolumn{1}{c|}{35.62$\pm$ 6.80} & \multicolumn{1}{c|}{11.93$\pm$ 5.45} & \multicolumn{1}{c|}{{\color[HTML]{002060} -23.69 (-67\%)}} & \multicolumn{1}{c|}{{\color[HTML]{002060} +0.64$\pm$ 0.49}} & \multicolumn{1}{c|}{0.25} \\ \cline{2-7} 
\multicolumn{1}{|c|}{} & \multicolumn{1}{c|}{MADNet} & \multicolumn{1}{l|}{36.03$\pm$ 20.20} & \multicolumn{1}{l|}{19.48$\pm$ 10.62} & \multicolumn{1}{c|}{{\color[HTML]{002060} -16.55 (-32\%)}} & \multicolumn{1}{c|}{{\color[HTML]{C00000} -0.14$\pm$ 0.59}} & \multicolumn{1}{c|}{0.88} \\ \cline{2-7} 
\multicolumn{1}{|c|}{} & \multicolumn{1}{c|}{HSM} & \multicolumn{1}{c|}{2.02$\pm$ 0.72} & \multicolumn{1}{c|}{3.03$\pm$ 1.03} & \multicolumn{1}{c|}{{\color[HTML]{C00000} +1.01 (+54\%)}} & \multicolumn{1}{c|}{{\color[HTML]{C00000} -1.00$\pm$ 0.00}} & \multicolumn{1}{c|}{1.00} \\ \cline{2-7} 
\multicolumn{1}{|c|}{} & \multicolumn{1}{c|}{GwcNet} & \multicolumn{1}{c|}{2.22$\pm$ 0.85} & \multicolumn{1}{c|}{2.76$\pm$ 0.80} & \multicolumn{1}{c|}{{\color[HTML]{C00000} +0.54 (+32\%)}} & \multicolumn{1}{c|}{{\color[HTML]{C00000} -0.96$\pm$ 0.05}} & \multicolumn{1}{c|}{1.00} \\ \cline{2-7} 
\multicolumn{1}{|c|}{\multirow{-5}{*}{SERV-CT}} & \multicolumn{1}{c|}{STTR} & \multicolumn{1}{c|}{2.04$\pm$ 0.76} & \multicolumn{1}{c|}{4.21$\pm$ 1.96} & \multicolumn{1}{c|}{{\color[HTML]{C00000} +2.16 (+141\%)}} & \multicolumn{1}{c|}{{\color[HTML]{C00000} -0.18$\pm$ 0.10}} & \multicolumn{1}{c|}{1.00} \\ \hline
\multicolumn{1}{l}{} & \multicolumn{1}{l}{} & \multicolumn{1}{l}{} & \multicolumn{1}{l}{} & \multicolumn{1}{l}{} &  &  \\ \hline 
\multicolumn{1}{|c|}{} & \multicolumn{1}{c|}{MonoDepth2} & \multicolumn{1}{c|}{14.54$\pm$ 4.11} & \multicolumn{1}{c|}{10.95$\pm$ 5.35} & \multicolumn{1}{c|}{{\color[HTML]{002060} -3.58 (-25\%)}} & \multicolumn{1}{c|}{{\color[HTML]{002060} +0.48$\pm$ 0.72}} & \multicolumn{1}{c|}{0.30} \\ \cline{2-7} 
\multicolumn{1}{|c|}{} & \multicolumn{1}{c|}{MADNet} & \multicolumn{1}{l|}{27.49$\pm$ 13.09} & \multicolumn{1}{c|}{17.54$\pm$ 6.23} & \multicolumn{1}{c|}{{\color[HTML]{002060} -9.95 (-25\%)}} & \multicolumn{1}{c|}{{\color[HTML]{C00000} +0.16$\pm$ 0.52}} & \multicolumn{1}{c|}{0.60} \\ \cline{2-7} 
\multicolumn{1}{|c|}{} & \multicolumn{1}{c|}{HSM} & \multicolumn{1}{c|}{2.05$\pm$ 0.90} & \multicolumn{1}{c|}{1.46$\pm$ 0.66} & \multicolumn{1}{c|}{{\color[HTML]{002060} -0.59 (-26\%)}} & \multicolumn{1}{c|}{{\color[HTML]{002060} +0.68$\pm$ 0.55}} & \multicolumn{1}{c|}{0.20} \\ \cline{2-7} 
\multicolumn{1}{|c|}{} & \multicolumn{1}{c|}{GwcNet} & \multicolumn{1}{c|}{1.00$\pm$ 0.37} & \multicolumn{1}{c|}{0.83$\pm$ 0.26} & \multicolumn{1}{c|}{{\color[HTML]{002060} -0.17 (-14\%)}} & \multicolumn{1}{c|}{{\color[HTML]{C00000} +0.19$\pm$ 0.59}} & \multicolumn{1}{c|}{0.60} \\ \cline{2-7} 
\multicolumn{1}{|c|}{\multirow{-5}{*}{Middlebury2014}} & \multicolumn{1}{c|}{STTR} & \multicolumn{1}{c|}{1.66$\pm$ 2.60} & \multicolumn{1}{c|}{1.34$\pm$ 1.24} & \multicolumn{1}{c|}{{\color[HTML]{002060} -0.33 (-5\%)}} & \multicolumn{1}{c|}{{\color[HTML]{C00000} -0.57$\pm$ 0.65}} & \multicolumn{1}{c|}{0.87} \\ \hline
\multicolumn{1}{l}{} & \textbf{} & \textbf{} & \textbf{} & \textbf{} &  &  \\ \hline
\multicolumn{1}{|c|}{} & \multicolumn{1}{c|}{MonoDepth2} & \multicolumn{1}{c|}{12.08$\pm$ 6.65} & \multicolumn{1}{c|}{3.84$\pm$ 4.44} & \multicolumn{1}{c|}{{\color[HTML]{002060} -8.24 (-73\%)}} & \multicolumn{1}{c|}{{\color[HTML]{002060} +0.85$\pm$ 0.21}} & \multicolumn{1}{c|}{0.04} \\ \cline{2-7} 
\multicolumn{1}{|c|}{} & \multicolumn{1}{c|}{MADNet} & \multicolumn{1}{c|}{26.98$\pm$ 11.62} & \multicolumn{1}{c|}{17.37$\pm$ 5.49} & \multicolumn{1}{c|}{{\color[HTML]{002060} -9.62 (-30\%)}} & \multicolumn{1}{c|}{{\color[HTML]{C00000} +0.36$\pm$ 0.40}} & \multicolumn{1}{c|}{0.44} \\ \cline{2-7} 
\multicolumn{1}{|c|}{} & \multicolumn{1}{c|}{HSM} & \multicolumn{1}{c|}{1.74$\pm$ 0.79} & \multicolumn{1}{c|}{1.50$\pm$ 0.51} & \multicolumn{1}{c|}{{\color[HTML]{002060} -0.24 (-10\%)}} & \multicolumn{1}{c|}{{\color[HTML]{C00000} +0.26$\pm$ 0.59}} & \multicolumn{1}{c|}{0.48} \\ \cline{2-7} 
\multicolumn{1}{|c|}{} & \multicolumn{1}{c|}{GwcNet} & \multicolumn{1}{c|}{1.30$\pm$ 0.46} & \multicolumn{1}{c|}{1.40$\pm$ 0.45} & \multicolumn{1}{c|}{{\color[HTML]{C00000} +0.10 (+10\%)}} & \multicolumn{1}{c|}{{\color[HTML]{C00000} -0.72$\pm$ 0.28}} & \multicolumn{1}{c|}{0.63} \\ \cline{2-7} 
\multicolumn{1}{|c|}{\multirow{-5}{*}{New Tsukuba}} & \multicolumn{1}{c|}{STTR} & \multicolumn{1}{c|}{1.39$\pm$ 0.48} & \multicolumn{1}{c|}{19.93$\pm$ 10.42} & \multicolumn{1}{c|}{{\color[HTML]{C00000} +18.54 (+1346\%)}} & \multicolumn{1}{c|}{{\color[HTML]{C00000} -0.94$\pm$ 0.08}} & \multicolumn{1}{c|}{1.00} \\ \hline
\multicolumn{1}{l}{} & \multicolumn{1}{c}{} & \multicolumn{1}{c}{} & \multicolumn{1}{c}{} & \multicolumn{1}{c|}{} & \multicolumn{1}{c|}{Average} & \multicolumn{1}{c|}{0.61} \\ \cline{6-7} 
\end{tabular}%
}
\end{table}

\subsection{Effect of Regularization} 
Since we observe diverging phenomenon during refinement, we further ablate the training with different regularization setting to ensure that it is not caused by a specific regularization. We conduct four experiments for each network: baseline without any regularization, with smoothness regularization only (denoted as SMTH), with multi-scale regularization only (denoted as MS), and with both regularization. We have selected SERV-CT and Middlebury2014 dataset for this experiment since it has \textit{dense} ground truth (allowing us to inspect every pixel) and distinct scene characteristics (medical and indoor). The quantitative result is summarized in \autoref{tab:reg}. 

We find that the commonly used regularization techniques do not always improve the final EPE performance. For example, for HSM/GwcNet/STTR trained on SERV-CT, the final EPE became worse that initial EPE after training ($\Delta$ column is {\color[HTML]{C00000} \textbf{red}}), regardless of the regularization techniques used. While it is commonly believed that using both regularizations gives the best result, we also find sometimes removing smoothness regularization leads to better result (MonoDepth2/MADNet/ trained on SERV-CT and MonoDepth2/GwcNet trained on Middlebury2014) and sometimes removing muti-scale regularization leads to better result (HSM trained on SERV-CT). These findings suggest that regularization techniques do not fundamentally solve the divergence issue and can lead to inconsistent result depending on the network/dataset.

\begin{table}[htpb]
\centering
\caption{Training result on two datasets with different types of regularization. MS: multi-scale regularization. SMTH: smoothness regularization. $\Delta$: changes between final and initial EPE ({\color[HTML]{002060} \textbf{blue}} indicates EPE decreases; {\color[HTML]{C00000} \textbf{red}} indicates EPE increases). $r_s$: Spearman's Rank Coefficient ({\color[HTML]{002060} \textbf{blue}} indicates average above 0.39; {\color[HTML]{C00000} \textbf{red}} indicates average below 0.39). Divergence: fraction of diverging training instances observed based on Definition 1.}
\label{tab:reg}
\resizebox{0.85\textwidth}{!}{%
\begin{tabular}{ccccccccc}
\hline
\multicolumn{1}{|c|}{} & \multicolumn{1}{c|}{} & \multicolumn{2}{c|}{\textbf{Regularization}} & \multicolumn{3}{c|}{\textbf{EPE}} & \multicolumn{1}{c|}{} & \multicolumn{1}{c|}{} \\ \cline{3-7}
\multicolumn{1}{|c|}{\multirow{-2}{*}{\textbf{Dataset}}} & \multicolumn{1}{c|}{\multirow{-2}{*}{\textbf{Network}}} & \multicolumn{1}{c|}{\textbf{MS}} & \multicolumn{1}{c|}{\textbf{SMTH}} & \multicolumn{1}{c|}{\textbf{initial}} & \multicolumn{1}{c|}{\textbf{Final}} & \multicolumn{1}{c|}{\textbf{$\Delta$}} & \multicolumn{1}{c|}{\multirow{-2}{*}{\textbf{$r_s$}}} & \multicolumn{1}{c|}{\multirow{-2}{*}{\textbf{Divergence}}} \\ \hline
\multicolumn{1}{|c|}{} & \multicolumn{1}{c|}{} & \multicolumn{1}{c|}{} & \multicolumn{1}{c|}{} & \multicolumn{1}{c|}{} & \multicolumn{1}{c|}{16.86 $\pm$ 8.07} & \multicolumn{1}{c|}{{\color[HTML]{002060} -19.05 (-52\%)}} & \multicolumn{1}{c|}{{\color[HTML]{002060} +0.59$\pm$ 0.42}} & \multicolumn{1}{c|}{0.25} \\ \cline{3-4} \cline{6-9} 
\multicolumn{1}{|c|}{} & \multicolumn{1}{c|}{} & \multicolumn{1}{c|}{} & \multicolumn{1}{c|}{\checkmark} & \multicolumn{1}{c|}{} & \multicolumn{1}{c|}{17.02 $\pm$ 7.93} & \multicolumn{1}{c|}{{\color[HTML]{002060} -18.90 (-52\%)}} & \multicolumn{1}{c|}{{\color[HTML]{002060} +0.54$\pm$ 0.43}} & \multicolumn{1}{c|}{0.31} \\ \cline{3-4} \cline{6-9} 
\multicolumn{1}{|c|}{} & \multicolumn{1}{c|}{} & \multicolumn{1}{c|}{\checkmark} & \multicolumn{1}{c|}{} & \multicolumn{1}{c|}{} & \multicolumn{1}{c|}{\textbf{16.81} $\pm$ 8.14} & \multicolumn{1}{c|}{{\color[HTML]{002060} -19.11 (-52\%)}} & \multicolumn{1}{c|}{{\color[HTML]{002060} +0.53$\pm$ 0.41}} & \multicolumn{1}{c|}{0.31} \\ \cline{3-4} \cline{6-9} 
\multicolumn{1}{|c|}{} & \multicolumn{1}{c|}{\multirow{-4}{*}{MonoDepth2}} & \multicolumn{1}{c|}{\checkmark} & \multicolumn{1}{c|}{\checkmark} & \multicolumn{1}{c|}{\multirow{-4}{*}{35.92 $\pm$ 9.93}} & \multicolumn{1}{c|}{17.00 $\pm$ 8.08} & \multicolumn{1}{c|}{{\color[HTML]{002060} -18.92 (-52\%)}} & \multicolumn{1}{c|}{{\color[HTML]{002060} +0.47$\pm$ 0.44}} & \multicolumn{1}{c|}{0.38} \\ \hhline{~========}
\multicolumn{1}{|c|}{} & \multicolumn{1}{c|}{} & \multicolumn{1}{c|}{} & \multicolumn{1}{c|}{} & \multicolumn{1}{c|}{} & \multicolumn{1}{c|}{8.07 $\pm$ 5.48} & \multicolumn{1}{c|}{{\color[HTML]{002060} -15.67 (-68\%)}} & \multicolumn{1}{c|}{{\color[HTML]{002060} +0.65$\pm$ 0.32}} & \multicolumn{1}{c|}{0.12} \\ \cline{3-4} \cline{6-9} 
\multicolumn{1}{|c|}{} & \multicolumn{1}{c|}{} & \multicolumn{1}{c|}{} & \multicolumn{1}{c|}{\checkmark} & \multicolumn{1}{c|}{} & \multicolumn{1}{c|}{8.06 $\pm$ 5.67} & \multicolumn{1}{c|}{{\color[HTML]{002060} -15.68 (-68\%)}} & \multicolumn{1}{c|}{{\color[HTML]{002060} +0.69$\pm$ 0.28}} & \multicolumn{1}{c|}{0.12} \\ \cline{3-4} \cline{6-9} 
\multicolumn{1}{|c|}{} & \multicolumn{1}{c|}{} & \multicolumn{1}{c|}{\checkmark} & \multicolumn{1}{c|}{} & \multicolumn{1}{c|}{} & \multicolumn{1}{c|}{\textbf{8.05} $\pm$ 5.58} & \multicolumn{1}{c|}{{\color[HTML]{002060} -15.69 (-68\%)}} & \multicolumn{1}{c|}{{\color[HTML]{002060} +0.66$\pm$ 0.32}} & \multicolumn{1}{c|}{0.12} \\ \cline{3-4} \cline{6-9} 
\multicolumn{1}{|c|}{} & \multicolumn{1}{c|}{\multirow{-4}{*}{MADNet}} & \multicolumn{1}{c|}{\checkmark} & \multicolumn{1}{c|}{\checkmark} & \multicolumn{1}{c|}{\multirow{-4}{*}{23.74 $\pm$ 9.68}} & \multicolumn{1}{c|}{7.97 $\pm$ 5.38} & \multicolumn{1}{c|}{{\color[HTML]{002060} -15.77 (-68\%)}} & \multicolumn{1}{c|}{{\color[HTML]{002060} +0.60$\pm$ 0.32}} & \multicolumn{1}{c|}{0.12} \\ \hhline{~========}
\multicolumn{1}{|c|}{} & \multicolumn{1}{c|}{} & \multicolumn{1}{c|}{} & \multicolumn{1}{c|}{} & \multicolumn{1}{c|}{} & \multicolumn{1}{c|}{4.01 $\pm$ 2.29} & \multicolumn{1}{c|}{{\color[HTML]{C00000} +1.88 (+93\%)}} & \multicolumn{1}{c|}{{\color[HTML]{C00000} -0.73$\pm$ 0.48}} & \multicolumn{1}{c|}{0.88} \\ \cline{3-4} \cline{6-9} 
\multicolumn{1}{|c|}{} & \multicolumn{1}{c|}{} & \multicolumn{1}{c|}{} & \multicolumn{1}{c|}{\checkmark} & \multicolumn{1}{c|}{} & \multicolumn{1}{c|}{\textbf{3.52} $\pm$ 1.99} & \multicolumn{1}{c|}{{\color[HTML]{C00000} +1.39 (+68\%)}} & \multicolumn{1}{c|}{{\color[HTML]{C00000} -0.55$\pm$ 0.61}} & \multicolumn{1}{c|}{0.88} \\ \cline{3-4} \cline{6-9} 
\multicolumn{1}{|c|}{} & \multicolumn{1}{c|}{} & \multicolumn{1}{c|}{\checkmark} & \multicolumn{1}{c|}{} & \multicolumn{1}{c|}{} & \multicolumn{1}{c|}{4.02 $\pm$ 2.53} & \multicolumn{1}{c|}{{\color[HTML]{C00000} +1.89 (+92\%)}} & \multicolumn{1}{c|}{{\color[HTML]{C00000} -0.65$\pm$ 0.45}} & \multicolumn{1}{c|}{0.94} \\ \cline{3-4} \cline{6-9} 
\multicolumn{1}{|c|}{} & \multicolumn{1}{c|}{\multirow{-4}{*}{HSM}} & \multicolumn{1}{c|}{\checkmark} & \multicolumn{1}{c|}{\checkmark} & \multicolumn{1}{c|}{\multirow{-4}{*}{2.13 $\pm$   0.73}} & \multicolumn{1}{c|}{3.58 $\pm$ 2.17} & \multicolumn{1}{c|}{{\color[HTML]{C00000} +1.44 (+70\%)}} & \multicolumn{1}{c|}{{\color[HTML]{C00000} -0.49$\pm$ 0.58}} & \multicolumn{1}{c|}{0.75} \\ \hhline{~========}
\multicolumn{1}{|c|}{} & \multicolumn{1}{c|}{} & \multicolumn{1}{c|}{} & \multicolumn{1}{c|}{} & \multicolumn{1}{c|}{} & \multicolumn{1}{c|}{5.33$\pm$ 1.58} & \multicolumn{1}{c|}{{\color[HTML]{C00000} +2.98 (+147\%)}} & \multicolumn{1}{c|}{{\color[HTML]{C00000} -0.99$\pm$ 0.01}} & \multicolumn{1}{c|}{1.00} \\ \cline{3-4} \cline{6-9} 
\multicolumn{1}{|c|}{} & \multicolumn{1}{c|}{} & \multicolumn{1}{c|}{} & \multicolumn{1}{c|}{\checkmark} & \multicolumn{1}{c|}{} & \multicolumn{1}{c|}{3.10$\pm$ 1.01} & \multicolumn{1}{c|}{{\color[HTML]{C00000} +0.83 (+42\%)}} & \multicolumn{1}{c|}{{\color[HTML]{C00000} -0.96$\pm$ 0.09}} & \multicolumn{1}{c|}{1.00} \\ \cline{3-4} \cline{6-9} 
\multicolumn{1}{|c|}{} & \multicolumn{1}{c|}{} & \multicolumn{1}{c|}{\checkmark} & \multicolumn{1}{c|}{} & \multicolumn{1}{c|}{} & \multicolumn{1}{c|}{5.41$\pm$ 1.78} & \multicolumn{1}{c|}{{\color[HTML]{C00000} +2.90 (+139\%)}} & \multicolumn{1}{c|}{{\color[HTML]{C00000} -1.00$\pm$ 0.00}} & \multicolumn{1}{c|}{1.00} \\ \cline{3-4} \cline{6-9} 
\multicolumn{1}{|c|}{} & \multicolumn{1}{c|}{\multirow{-4}{*}{GwcNet}} & \multicolumn{1}{c|}{\checkmark} & \multicolumn{1}{c|}{\checkmark} & \multicolumn{1}{c|}{\multirow{-4}{*}{2.22$\pm$ 0.85}} & \multicolumn{1}{c|}{\textbf{2.76} $\pm$ 0.80} & \multicolumn{1}{c|}{{\color[HTML]{C00000} +0.54 (+32\%)}} & \multicolumn{1}{c|}{{\color[HTML]{C00000} -0.96$\pm$ 0.05}} & \multicolumn{1}{c|}{1.00} \\ \hhline{~========}
\multicolumn{1}{|c|}{} & \multicolumn{1}{c|}{} & \multicolumn{1}{c|}{} & \multicolumn{1}{c|}{} & \multicolumn{1}{c|}{} & \multicolumn{1}{c|}{5.42 $\pm$ 3.50} & \multicolumn{1}{c|}{{\color[HTML]{C00000} +1.98 (+57\%)}} & \multicolumn{1}{c|}{{\color[HTML]{C00000} -0.60$\pm$ 0.60}} & \multicolumn{1}{c|}{0.81} \\ \cline{3-4} \cline{6-9} 
\multicolumn{1}{|c|}{} & \multicolumn{1}{c|}{} & \multicolumn{1}{c|}{} & \multicolumn{1}{c|}{\checkmark} & \multicolumn{1}{c|}{} & \multicolumn{1}{c|}{4.38 $\pm$ 2.81} & \multicolumn{1}{c|}{{\color[HTML]{C00000} +0.93 (+29\%)}} & \multicolumn{1}{c|}{{\color[HTML]{C00000} -0.54$\pm$ 0.58}} & \multicolumn{1}{c|}{0.88} \\ \cline{3-4} \cline{6-9} 
\multicolumn{1}{|c|}{} & \multicolumn{1}{c|}{} & \multicolumn{1}{c|}{\checkmark} & \multicolumn{1}{c|}{} & \multicolumn{1}{c|}{} & \multicolumn{1}{c|}{4.46 $\pm$ 2.66} & \multicolumn{1}{c|}{{\color[HTML]{C00000} +1.02 (+32\%)}} & \multicolumn{1}{c|}{{\color[HTML]{C00000} -0.70$\pm$ 0.36}} & \multicolumn{1}{c|}{0.94} \\ \cline{3-4} \cline{6-9} 
\multicolumn{1}{|c|}{\multirow{-20}{*}{SERV-CT}} & \multicolumn{1}{c|}{\multirow{-4}{*}{STTR}} & \multicolumn{1}{c|}{\checkmark} & \multicolumn{1}{c|}{\checkmark} & \multicolumn{1}{c|}{\multirow{-4}{*}{3.44 $\pm$ 2.02}} & \multicolumn{1}{c|}{\textbf{3.48} $\pm$ 1.83} & \multicolumn{1}{c|}{{\color[HTML]{C00000} +0.03 (+5\%)}} & \multicolumn{1}{c|}{{\color[HTML]{C00000} -0.39$\pm$ 0.53}} & \multicolumn{1}{c|}{0.81} \\ \hline
 &  &  &  &  &  & {\color[HTML]{C00000} } & {\color[HTML]{C00000} } &  \\ \hline
\multicolumn{1}{|c|}{} & \multicolumn{1}{c|}{} & \multicolumn{1}{c|}{} & \multicolumn{1}{c|}{} & \multicolumn{1}{c|}{} & \multicolumn{1}{c|}{10.95$\pm$ 5.41} & \multicolumn{1}{c|}{{\color[HTML]{002060} -3.75 (-26\%)}} & \multicolumn{1}{c|}{{\color[HTML]{002060} +0.39$\pm$ 0.87}} & \multicolumn{1}{c|}{0.30} \\ \cline{3-4} \cline{6-9} 
\multicolumn{1}{|c|}{} & \multicolumn{1}{c|}{} & \multicolumn{1}{c|}{} & \multicolumn{1}{c|}{\checkmark} & \multicolumn{1}{c|}{} & \multicolumn{1}{c|}{11.07$\pm$ 5.16} & \multicolumn{1}{c|}{{\color[HTML]{002060} -3.58 (-25\%)}} & \multicolumn{1}{c|}{{\color[HTML]{002060} +0.42$\pm$ 0.83}} & \multicolumn{1}{c|}{0.30} \\ \cline{3-4} \cline{6-9} 
\multicolumn{1}{|c|}{} & \multicolumn{1}{c|}{} & \multicolumn{1}{c|}{\checkmark} & \multicolumn{1}{c|}{} & \multicolumn{1}{c|}{} & \multicolumn{1}{c|}{\textbf{10.88} $\pm$ 5.37} & \multicolumn{1}{c|}{{\color[HTML]{002060} -3.70 (-26\%)}} & \multicolumn{1}{c|}{{\color[HTML]{002060} +0.45$\pm$ 0.76}} & \multicolumn{1}{c|}{0.30} \\ \cline{3-4} \cline{6-9} 
\multicolumn{1}{|c|}{} & \multicolumn{1}{c|}{\multirow{-4}{*}{MonoDepth2}} & \multicolumn{1}{c|}{\checkmark} & \multicolumn{1}{c|}{\checkmark} & \multicolumn{1}{c|}{\multirow{-4}{*}{14.54$\pm$   4.11}} & \multicolumn{1}{c|}{10.95$\pm$ 5.35} & \multicolumn{1}{c|}{{\color[HTML]{002060} -3.58 (-25\%)}} & \multicolumn{1}{c|}{{\color[HTML]{002060} +0.48$\pm$ 0.72}} & \multicolumn{1}{c|}{0.30} \\ \hhline{~========}
\multicolumn{1}{|c|}{} & \multicolumn{1}{c|}{} & \multicolumn{1}{c|}{} & \multicolumn{1}{c|}{} & \multicolumn{1}{c|}{} & \multicolumn{1}{c|}{\textbf{17.01} $\pm$ 6.07} & \multicolumn{1}{c|}{{\color[HTML]{002060} -11.01 (-29\%)}} & \multicolumn{1}{c|}{{\color[HTML]{C00000} +0.20$\pm$ 0.56}} & \multicolumn{1}{c|}{0.40} \\ \cline{3-4} \cline{6-9} 
\multicolumn{1}{|c|}{} & \multicolumn{1}{c|}{} & \multicolumn{1}{c|}{} & \multicolumn{1}{c|}{\checkmark} & \multicolumn{1}{c|}{} & \multicolumn{1}{c|}{17.38$\pm$ 6.09} & \multicolumn{1}{c|}{{\color[HTML]{002060} -10.44 (-26\%)}} & \multicolumn{1}{c|}{{\color[HTML]{C00000} +0.29$\pm$ 0.51}} & \multicolumn{1}{c|}{0.30} \\ \cline{3-4} \cline{6-9} 
\multicolumn{1}{|c|}{} & \multicolumn{1}{c|}{} & \multicolumn{1}{c|}{\checkmark} & \multicolumn{1}{c|}{} & \multicolumn{1}{c|}{} & \multicolumn{1}{c|}{18.17$\pm$ 6.20} & \multicolumn{1}{c|}{{\color[HTML]{002060} -9.14 (-25\%)}} & \multicolumn{1}{c|}{{\color[HTML]{C00000} +0.15$\pm$ 0.52}} & \multicolumn{1}{c|}{0.30} \\ \cline{3-4} \cline{6-9} 
\multicolumn{1}{|c|}{} & \multicolumn{1}{c|}{\multirow{-4}{*}{MADNet}} & \multicolumn{1}{c|}{\checkmark} & \multicolumn{1}{c|}{\checkmark} & \multicolumn{1}{c|}{\multirow{-4}{*}{27.49$\pm$ 13.09}} & \multicolumn{1}{c|}{17.54$\pm$ 6.23} & \multicolumn{1}{c|}{{\color[HTML]{002060} -9.95 (-25\%)}} & \multicolumn{1}{c|}{{\color[HTML]{C00000} +0.16$\pm$ 0.52}} & \multicolumn{1}{c|}{0.40} \\ \hhline{~========}
\multicolumn{1}{|c|}{} & \multicolumn{1}{c|}{} & \multicolumn{1}{c|}{} & \multicolumn{1}{c|}{} & \multicolumn{1}{c|}{} & \multicolumn{1}{c|}{\textbf{1.24}$\pm$ 0.52} & \multicolumn{1}{c|}{{\color[HTML]{002060} -0.71 (-35\%)}} & \multicolumn{1}{c|}{{\color[HTML]{C00000} +0.20$\pm$ 0.56}} & \multicolumn{1}{c|}{0.40} \\ \cline{3-4} \cline{6-9} 
\multicolumn{1}{|c|}{} & \multicolumn{1}{c|}{} & \multicolumn{1}{c|}{} & \multicolumn{1}{c|}{\checkmark} & \multicolumn{1}{c|}{} & \multicolumn{1}{c|}{1.30$\pm$ 0.60} & \multicolumn{1}{c|}{{\color[HTML]{002060} -0.63 (-32\%)}} & \multicolumn{1}{c|}{{\color[HTML]{C00000} +0.29$\pm$ 0.51}} & \multicolumn{1}{c|}{0.30} \\ \cline{3-4} \cline{6-9} 
\multicolumn{1}{|c|}{} & \multicolumn{1}{c|}{} & \multicolumn{1}{c|}{\checkmark} & \multicolumn{1}{c|}{} & \multicolumn{1}{c|}{} & \multicolumn{1}{c|}{\textbf{1.24}$\pm$ 0.51} & \multicolumn{1}{c|}{{\color[HTML]{002060} -0.71 (-35\%)}} & \multicolumn{1}{c|}{{\color[HTML]{C00000} +0.15$\pm$ 0.52}} & \multicolumn{1}{c|}{0.30} \\ \cline{3-4} \cline{6-9} 
\multicolumn{1}{|c|}{} & \multicolumn{1}{c|}{\multirow{-4}{*}{HSM}} & \multicolumn{1}{c|}{\checkmark} & \multicolumn{1}{c|}{\checkmark} & \multicolumn{1}{c|}{\multirow{-4}{*}{2.05$\pm$   0.90}} & \multicolumn{1}{c|}{1.27$\pm$ 0.58} & \multicolumn{1}{c|}{{\color[HTML]{002060} -0.68 (-34\%)}} & \multicolumn{1}{c|}{{\color[HTML]{C00000} +0.16$\pm$ 0.52}} & \multicolumn{1}{c|}{0.40} \\ \hhline{~========}
\multicolumn{1}{|c|}{} & \multicolumn{1}{c|}{} & \multicolumn{1}{c|}{} & \multicolumn{1}{c|}{} & \multicolumn{1}{c|}{} & \multicolumn{1}{c|}{1.04$\pm$ 0.94} & \multicolumn{1}{c|}{{\color[HTML]{C00000} +0.09 (+6\%)}} & \multicolumn{1}{c|}{{\color[HTML]{C00000} +0.24$\pm$ 0.65}} & \multicolumn{1}{c|}{0.40} \\ \cline{3-4} \cline{6-9} 
\multicolumn{1}{|c|}{} & \multicolumn{1}{c|}{} & \multicolumn{1}{c|}{} & \multicolumn{1}{c|}{\checkmark} & \multicolumn{1}{c|}{} & \multicolumn{1}{c|}{0.78$\pm$ 0.21} & \multicolumn{1}{c|}{{\color[HTML]{002060} -0.17 (-14\%)}} & \multicolumn{1}{c|}{{\color[HTML]{C00000} -0.10$\pm$ 0.74}} & \multicolumn{1}{c|}{0.50} \\ \cline{3-4} \cline{6-9} 
\multicolumn{1}{|c|}{} & \multicolumn{1}{c|}{} & \multicolumn{1}{c|}{\checkmark} & \multicolumn{1}{c|}{} & \multicolumn{1}{c|}{} & \multicolumn{1}{c|}{\textbf{0.77}$\pm$ 0.24} & \multicolumn{1}{c|}{{\color[HTML]{002060} -0.23 (-19\%)}} & \multicolumn{1}{c|}{{\color[HTML]{C00000} -0.01$\pm$ 0.67}} & \multicolumn{1}{c|}{0.50} \\ \cline{3-4} \cline{6-9} 
\multicolumn{1}{|c|}{} & \multicolumn{1}{c|}{\multirow{-4}{*}{GwcNet}} & \multicolumn{1}{c|}{\checkmark} & \multicolumn{1}{c|}{\checkmark} & \multicolumn{1}{c|}{\multirow{-4}{*}{1.00$\pm$   0.37}} & \multicolumn{1}{c|}{0.83$\pm$ 0.26} & \multicolumn{1}{c|}{{\color[HTML]{002060} -0.17 (-14\%)}} & \multicolumn{1}{c|}{{\color[HTML]{C00000} +0.19$\pm$ 0.59}} & \multicolumn{1}{c|}{0.60} \\ \hhline{~========}
\multicolumn{1}{|c|}{} & \multicolumn{1}{c|}{} & \multicolumn{1}{c|}{} & \multicolumn{1}{c|}{} & \multicolumn{1}{c|}{} & \multicolumn{1}{c|}{2.34$\pm$ 4.57} & \multicolumn{1}{c|}{{\color[HTML]{C00000} +0.69 (+20\%)}} & \multicolumn{1}{c|}{{\color[HTML]{C00000} -0.74$\pm$ 0.53}} & \multicolumn{1}{c|}{0.87} \\ \cline{3-4} \cline{6-9} 
\multicolumn{1}{|c|}{} & \multicolumn{1}{c|}{} & \multicolumn{1}{c|}{} & \multicolumn{1}{c|}{\checkmark} & \multicolumn{1}{c|}{} & \multicolumn{1}{c|}{2.48$\pm$ 5.36} & \multicolumn{1}{c|}{{\color[HTML]{C00000} +0.77 (+14\%)}} & \multicolumn{1}{c|}{{\color[HTML]{C00000} -0.64$\pm$ 0.64}} & \multicolumn{1}{c|}{0.87} \\ \cline{3-4} \cline{6-9} 
\multicolumn{1}{|c|}{} & \multicolumn{1}{c|}{} & \multicolumn{1}{c|}{\checkmark} & \multicolumn{1}{c|}{} & \multicolumn{1}{c|}{} & \multicolumn{1}{c|}{2.05$\pm$ 3.58} & \multicolumn{1}{c|}{{\color[HTML]{C00000} +0.39 (+18\%)}} & \multicolumn{1}{c|}{{\color[HTML]{C00000} -0.63$\pm$ 0.57}} & \multicolumn{1}{c|}{0.80} \\ \cline{3-4} \cline{6-9} 
\multicolumn{1}{|c|}{\multirow{-20}{*}{Middlebury2014}} & \multicolumn{1}{c|}{\multirow{-4}{*}{STTR}} & \multicolumn{1}{c|}{\checkmark} & \multicolumn{1}{c|}{\checkmark} & \multicolumn{1}{c|}{\multirow{-4}{*}{1.66$\pm$ 2.60}} & \multicolumn{1}{c|}{\textbf{1.34}$\pm$ 1.23} & \multicolumn{1}{c|}{{\color[HTML]{002060} -0.31 (-5\%)}} & \multicolumn{1}{c|}{{\color[HTML]{C00000} -0.57$\pm$ 0.65}} & \multicolumn{1}{c|}{0.87} \\ \hline
\end{tabular}%
}
\end{table}

\section{Sensitivity Analysis of Thresholds for Definition 1}
\begin{figure}[htpb]
    \centering
    \subfloat[$r_s=1.00, \epsilon=0.0$]{\includegraphics[width=0.3\textwidth]{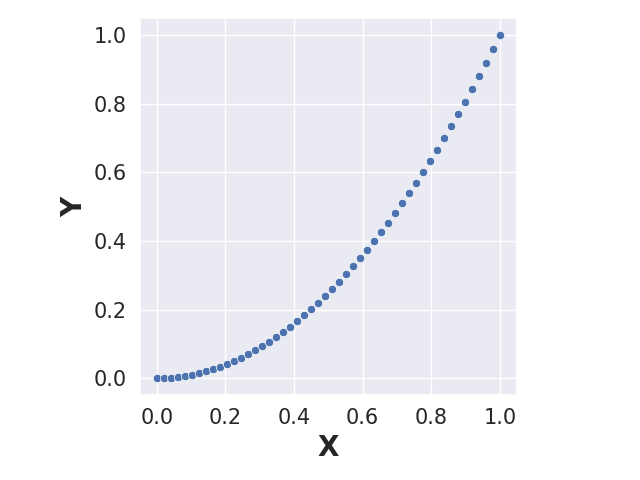}}
    \subfloat[$r_s=0.87, \epsilon=0.5$]{\includegraphics[width=0.3\textwidth]{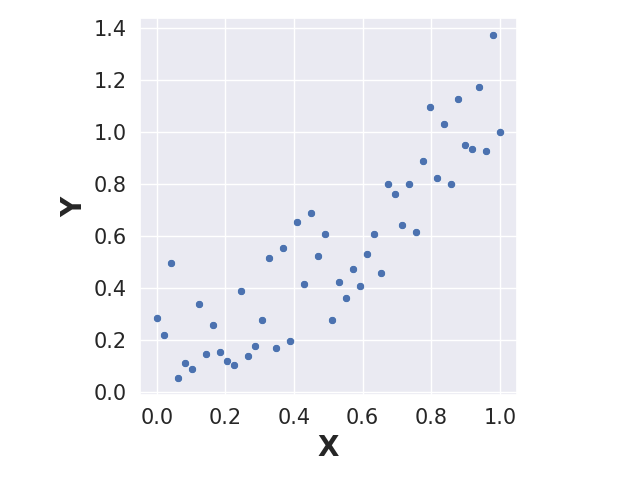}}
    \subfloat[$r_s=0.73, \epsilon=1.0$]{\includegraphics[width=0.3\textwidth]{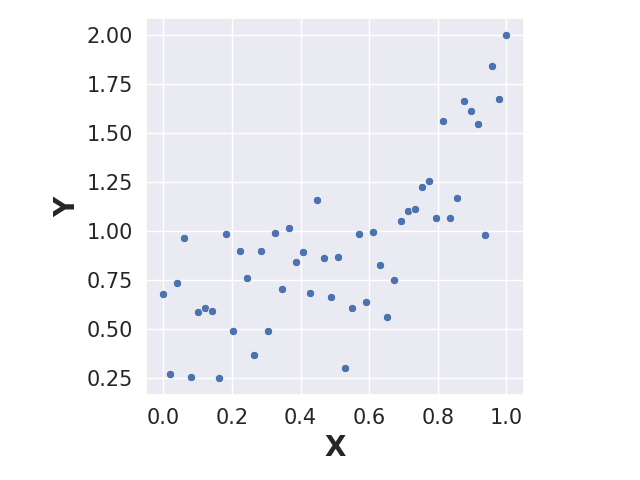}}
    
    \subfloat[$r_s=0.48, \epsilon=1.5$]{\includegraphics[width=0.3\textwidth]{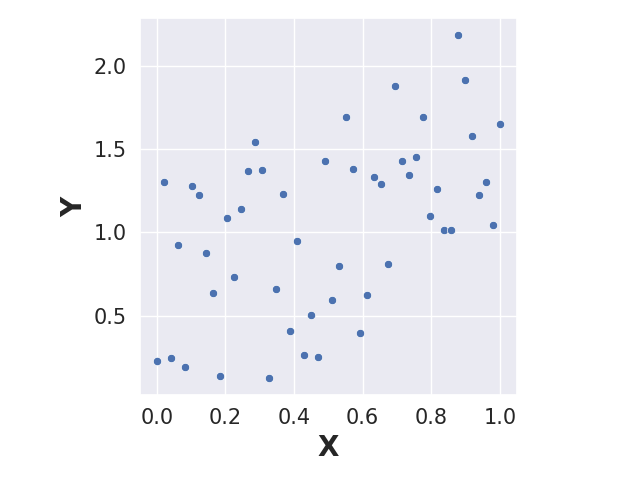}}
    \subfloat[$r_s=0.39,\epsilon=2.0$]{\includegraphics[width=0.3\textwidth]{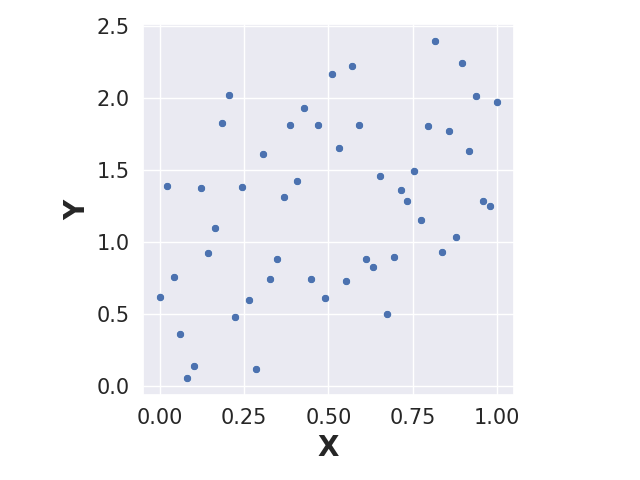}}
    \subfloat[$r_s=0.29,\epsilon=2.5$]{\includegraphics[width=0.3\textwidth]{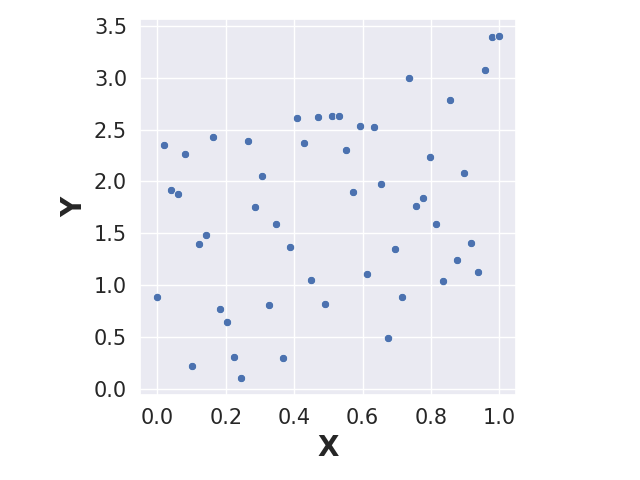}}
    \caption{Simulation result of $y=x^2$ with increasing noise and decreasing Spearman's Rank Coefficient $r_s$.}
    \label{fig:spearman_sensitivity}
\end{figure}

In \autoref{fig:spearman_sensitivity}, we generate a set of simulation result where the original variables are perfectly correlated with $y=x^2$, with $x=\{0.0, 0.02, 0.04, ...,1.0\}$ at 0.02 interval. We add increasing corruption that is randomly sampled from the range [0.0,\, $\epsilon$] to $y$ to qualitatively illustrate the how Spearman's Rank coefficient $r_s$ changes as data change. As $\epsilon$ increased from 0.0 to 2.5, Spearman's Rank Coefficient decreases from 1.0 to 0.29 and the correlation between $x,y$ are becoming weaker.

One other well-established threshold of Spearman's Rank coefficient $r_s$ for weak correlation in the political and medical researches is 0.29 \citep{akoglu2018user}. We present result in Table 1 again but recomputed with the new threshold. The presented results only differ in two cases (MADNet trained on KITTI2015 and Middlebury2014) and average divergence case changes from 0.54 to 0.50, thus indicating our definition of divergence in Definition 1 is not highly sensitive to the selected threshold. 

\begin{table}[htpb]
\centering
\caption{Comparison of results with Spearman's Rank coefficient $r_s$ threshold being 0.39 and 0.29.}
\label{tab:my-table}
\resizebox{0.7\textwidth}{!}{%
\begin{tabular}{cccccc}
\cline{3-6}
\multicolumn{1}{l}{} & \multicolumn{1}{l|}{\textbf{}} & \multicolumn{2}{c|}{\textbf{Threshold=0.39}} & \multicolumn{2}{c|}{\textbf{Threshold=0.29}} \\ \hline
\multicolumn{1}{|c|}{\textbf{Dataset}} & \multicolumn{1}{c|}{\textbf{Network}} & \multicolumn{1}{c|}{\textbf{$r_s$}} & \multicolumn{1}{c|}{\textbf{Divergence}} & \multicolumn{1}{c|}{\textbf{$r_s$}} & \multicolumn{1}{c|}{\textbf{Divergence}} \\ \hline
\multicolumn{1}{|c|}{} & \multicolumn{1}{c|}{MonoDepth2} & \multicolumn{1}{c|}{{\color[HTML]{002060} +0.41$\pm$ 0.38}} & \multicolumn{1}{c|}{0.20} & \multicolumn{1}{c|}{{\color[HTML]{002060} +0.41$\pm$ 0.38}} & \multicolumn{1}{c|}{0.16} \\ \cline{2-6} 
\multicolumn{1}{|c|}{} & \multicolumn{1}{c|}{MADNet} & \multicolumn{1}{c|}{{\color[HTML]{C00000} +0.35$\pm$ 0.45}} & \multicolumn{1}{c|}{0.36} & \multicolumn{1}{c|}{{\color[HTML]{002060} +0.35$\pm$ 0.45}} & \multicolumn{1}{c|}{0.30} \\ \cline{2-6} 
\multicolumn{1}{|c|}{} & \multicolumn{1}{c|}{HSM} & \multicolumn{1}{c|}{{\color[HTML]{C00000} +0.04$\pm$ 0.60}} & \multicolumn{1}{c|}{0.48} & \multicolumn{1}{c|}{{\color[HTML]{C00000} +0.04$\pm$ 0.60}} & \multicolumn{1}{c|}{0.44} \\ \cline{2-6} 
\multicolumn{1}{|c|}{} & \multicolumn{1}{c|}{GwcNet} & \multicolumn{1}{c|}{{\color[HTML]{C00000} -0.41$\pm$ 0.58}} & \multicolumn{1}{c|}{0.82} & \multicolumn{1}{c|}{{\color[HTML]{C00000} -0.41$\pm$ 0.58}} & \multicolumn{1}{c|}{0.76} \\ \cline{2-6} 
\multicolumn{1}{|c|}{\multirow{-5}{*}{KITTI2015}} & \multicolumn{1}{c|}{STTR} & \multicolumn{1}{c|}{{\color[HTML]{C00000} -0.01$\pm$ 0.21}} & \multicolumn{1}{c|}{0.40} & \multicolumn{1}{c|}{{\color[HTML]{C00000} -0.01$\pm$ 0.21}} & \multicolumn{1}{c|}{0.38} \\ \hline
\multicolumn{1}{l}{} &  &  &  &  &  \\ \hline
\multicolumn{1}{|c|}{} & \multicolumn{1}{c|}{MonoDepth2} & \multicolumn{1}{c|}{{\color[HTML]{002060} +0.65$\pm$ 0.51}} & \multicolumn{1}{c|}{0.25} & \multicolumn{1}{c|}{{\color[HTML]{002060} +0.65$\pm$ 0.51}} & \multicolumn{1}{c|}{0.25} \\ \cline{2-6} 
\multicolumn{1}{|c|}{} & \multicolumn{1}{c|}{MADNet} & \multicolumn{1}{c|}{{\color[HTML]{C00000} +0.34$\pm$ 0.60}} & \multicolumn{1}{c|}{0.50} & \multicolumn{1}{c|}{{\color[HTML]{002060} +0.34$\pm$ 0.60}} & \multicolumn{1}{c|}{0.25} \\ \cline{2-6} 
\multicolumn{1}{|c|}{} & \multicolumn{1}{c|}{HSM} & \multicolumn{1}{c|}{{\color[HTML]{C00000} -0.29$\pm$ 0.67}} & \multicolumn{1}{c|}{0.62} & \multicolumn{1}{c|}{{\color[HTML]{C00000} -0.29$\pm$ 0.67}} & \multicolumn{1}{c|}{0.62} \\ \cline{2-6} 
\multicolumn{1}{|c|}{} & \multicolumn{1}{c|}{GwcNet} & \multicolumn{1}{c|}{{\color[HTML]{C00000} -0.58$\pm$ 0.56}} & \multicolumn{1}{c|}{0.88} & \multicolumn{1}{c|}{{\color[HTML]{C00000} -0.58$\pm$ 0.56}} & \multicolumn{1}{c|}{0.88} \\ \cline{2-6} 
\multicolumn{1}{|c|}{\multirow{-5}{*}{SERV-CT}} & \multicolumn{1}{c|}{STTR} & \multicolumn{1}{c|}{{\color[HTML]{C00000} -0.73$\pm$ 0.38}} & \multicolumn{1}{c|}{1.00} & \multicolumn{1}{c|}{{\color[HTML]{C00000} -0.73$\pm$ 0.38}} & \multicolumn{1}{c|}{1.00} \\ \hline
\multicolumn{1}{l}{} & \multicolumn{1}{l}{} &  &  &  &  \\ \hline
\multicolumn{1}{|c|}{} & \multicolumn{1}{c|}{MonoDepth2} & \multicolumn{1}{c|}{{\color[HTML]{C00000} -0.28$\pm$ 0.83}} & \multicolumn{1}{c|}{0.60} & \multicolumn{1}{c|}{{\color[HTML]{C00000} -0.28$\pm$ 0.83}} & \multicolumn{1}{c|}{0.60} \\ \cline{2-6} 
\multicolumn{1}{|c|}{} & \multicolumn{1}{c|}{MADNet} & \multicolumn{1}{c|}{{\color[HTML]{C00000} -0.29$\pm$ 0.51}} & \multicolumn{1}{c|}{0.60} & \multicolumn{1}{c|}{{\color[HTML]{C00000} -0.29$\pm$ 0.51}} & \multicolumn{1}{c|}{0.60} \\ \cline{2-6} 
\multicolumn{1}{|c|}{} & \multicolumn{1}{c|}{HSM} & \multicolumn{1}{c|}{{\color[HTML]{002060} +0.89$\pm$ 0.17}} & \multicolumn{1}{c|}{0.00} & \multicolumn{1}{c|}{{\color[HTML]{002060} +0.89$\pm$ 0.17}} & \multicolumn{1}{c|}{0.00} \\ \cline{2-6} 
\multicolumn{1}{|c|}{} & \multicolumn{1}{c|}{GwcNet} & \multicolumn{1}{c|}{{\color[HTML]{002060} +0.42$\pm$ 0.60}} & \multicolumn{1}{c|}{0.20} & \multicolumn{1}{c|}{{\color[HTML]{002060} +0.42$\pm$ 0.60}} & \multicolumn{1}{c|}{0.20} \\ \cline{2-6} 
\multicolumn{1}{|c|}{\multirow{-5}{*}{Middlebury2014}} & \multicolumn{1}{c|}{STTR} & \multicolumn{1}{c|}{{\color[HTML]{C00000} -0.58$\pm$ 0.65}} & \multicolumn{1}{c|}{0.87} & \multicolumn{1}{c|}{{\color[HTML]{C00000} -0.58$\pm$ 0.65}} & \multicolumn{1}{c|}{0.87} \\ \hline
\multicolumn{1}{l}{} & \textbf{} &  &  &  &  \\ \hline
\multicolumn{1}{|c|}{} & \multicolumn{1}{c|}{MonoDepth2} & \multicolumn{1}{c|}{{\color[HTML]{C00000} +0.26$\pm$ 0.38}} & \multicolumn{1}{c|}{0.38} & \multicolumn{1}{c|}{{\color[HTML]{C00000} +0.26$\pm$ 0.38}} & \multicolumn{1}{c|}{0.22} \\ \cline{2-6} 
\multicolumn{1}{|c|}{} & \multicolumn{1}{c|}{MADNet} & \multicolumn{1}{c|}{{\color[HTML]{C00000} 0.00$\pm$ 0.42}} & \multicolumn{1}{c|}{0.56} & \multicolumn{1}{c|}{{\color[HTML]{C00000} 0.00$\pm$ 0.42}} & \multicolumn{1}{c|}{0.48} \\ \cline{2-6} 
\multicolumn{1}{|c|}{} & \multicolumn{1}{c|}{HSM} & \multicolumn{1}{c|}{{\color[HTML]{C00000} +0.15$\pm$ 0.47}} & \multicolumn{1}{c|}{0.48} & \multicolumn{1}{c|}{{\color[HTML]{C00000} +0.15$\pm$ 0.47}} & \multicolumn{1}{c|}{0.40} \\ \cline{2-6} 
\multicolumn{1}{|c|}{} & \multicolumn{1}{c|}{GwcNet} & \multicolumn{1}{c|}{{\color[HTML]{C00000} -0.68$\pm$ 0.21}} & \multicolumn{1}{c|}{0.56} & \multicolumn{1}{c|}{{\color[HTML]{C00000} -0.68$\pm$ 0.21}} & \multicolumn{1}{c|}{0.52} \\ \cline{2-6} 
\multicolumn{1}{|c|}{\multirow{-5}{*}{Tsukuba}} & \multicolumn{1}{c|}{STTR} & \multicolumn{1}{c|}{{\color[HTML]{C00000} -0.89$\pm$ 0.16}} & \multicolumn{1}{c|}{1.00} & \multicolumn{1}{c|}{{\color[HTML]{C00000} -0.89$\pm$ 0.16}} & \multicolumn{1}{c|}{1.00} \\ \hline
\multicolumn{1}{c}{} & \multicolumn{1}{c|}{} & \multicolumn{1}{c|}{Average} & \multicolumn{1}{c|}{0.54} & \multicolumn{1}{c|}{Average} & \multicolumn{1}{c|}{0.50} \\ \cline{3-6}
\end{tabular}%
}
\end{table}

\section{Change of EPE with decreasing \syn}
\begin{figure}[htpb]
    \centering
    \subfloat[SERV-CT]{\includegraphics[height=0.35\textwidth]{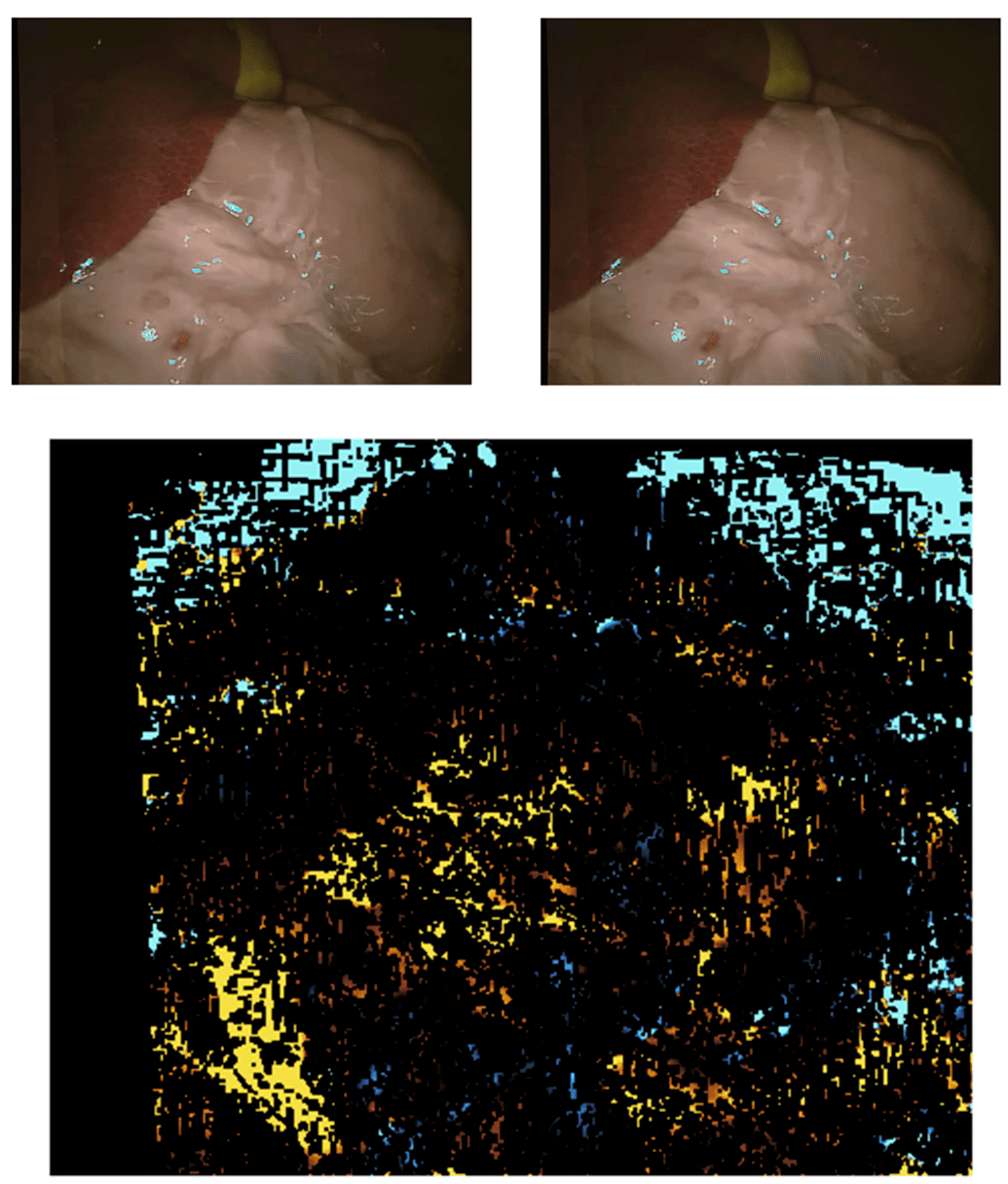}}
    \subfloat[New Tsukuba]{\includegraphics[height=0.35\textwidth]{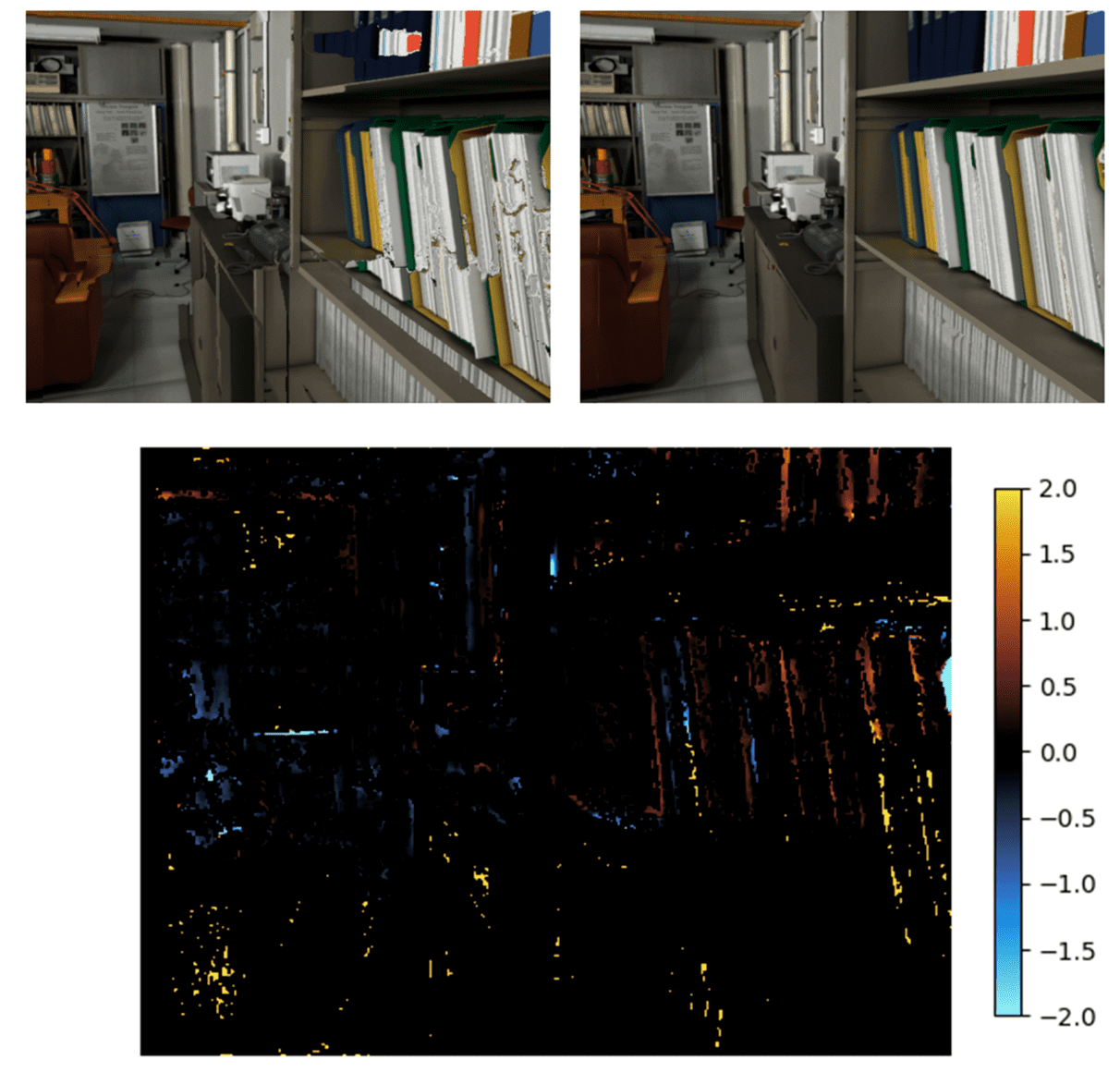}}
    \caption{Visualization of synthesized images and change of EPE at places where \syn~decreases (black are either occluded regions or where \syn~increases). Top left figure is the initial synthesized image, top right figure is the final synthesized image, and the bottom figure is the change of EPE.}
    \label{fig:local_comparison}
\end{figure}

We provide qualitative results of the change of EPE with decreasing \syn. We mask out the pixels that are occluded or with increasing \syn~as black color. As shown in \autoref{fig:local_comparison}, while some of the pixels have decreasing EPE (cooler color), many have increased EPE (warmer color). The increasing EPE are particularly evident at occlusion/object edges, textureless areas and specular regions as discussed in Section 4.4.

\section{Effect of Weighting Between $L_{SSIM}$ and $L_{L1}$}
As discussed in Section 4.3, the gradient w.r.t disparity estimation will be non-zero due to the residual error (where gradient is loss times learning rate), this means if there is a network with perfect depth prediction, it will diverge from this point. We examine if weightings between \ssim~and \lone~affect training behavior. We conduct three experiments, using the default weighting in Equation 1, using $L_{SSIM}$ alone, and using $L_{L1}$ alone. Experiments are conducted \textit{with} multi-scale and smoothness regularization. Quantitative result is summarized in \autoref{tab:weighting}.

We find that there are cases where using either loss or a combination of them alone will lead to worse EPE, such as HSM/GwcNet trained on SERV-CT ($\Delta$ column is {\color[HTML]{C00000} \textbf{red}}). We conclude that the divergent problem is not due to the weighting of \ssim~and \lone, therefore adjusting the weight will not solve the divergence issue. We also find that using \lone~alone is consistently worse than using \ssim~alone with the exception of MADNet trained on Middlebury2014, which justifies the larger weighting proposed for \ssim~in \citep{zhao2015loss,godard2017unsupervised}. Moreover, we find that sometimes using \ssim~alone is better, especially in dataset such as SERV-CT where many specular reflective regions are present.

\begin{table}[htpb]
\centering
\caption{Training result on two datasets with different weightings between \ssim~and \lone. $\Delta$: changes between final and initial EPE ({\color[HTML]{002060} \textbf{blue}} indicates EPE decreases; {\color[HTML]{C00000} \textbf{red}} indicates EPE increases). $r_s$: Spearman's Rank Coefficient ({\color[HTML]{002060} \textbf{blue}} indicates average above 0.39; {\color[HTML]{C00000} \textbf{red}} indicates average below 0.39). Divergence: fraction of diverging training instances observed based on Definition 1.}
\label{tab:weighting}
\resizebox{0.85\textwidth}{!}{%
\begin{tabular}{ccccccccc}
\hline
\multicolumn{1}{|c|}{} & \multicolumn{1}{c|}{} & \multicolumn{2}{c|}{\textbf{Weighting}} & \multicolumn{3}{c|}{\textbf{EPE}} & \multicolumn{1}{c|}{} & \multicolumn{1}{c|}{} \\ \cline{3-7}
\multicolumn{1}{|c|}{\multirow{-2}{*}{\textbf{Dataset}}} & \multicolumn{1}{c|}{\multirow{-2}{*}{\textbf{Network}}} & \multicolumn{1}{c|}{\textbf{\ssim}} & \multicolumn{1}{c|}{\textbf{\lone}} & \multicolumn{1}{c|}{\textbf{initial}} & \multicolumn{1}{c|}{\textbf{Final}} & \multicolumn{1}{c|}{\textbf{$\Delta$}} & \multicolumn{1}{c|}{\multirow{-2}{*}{\textbf{$r_s$}}} & \multicolumn{1}{c|}{\multirow{-2}{*}{\textbf{Divergence}}} \\ \hline
\multicolumn{1}{|c|}{} & \multicolumn{1}{c|}{} & \multicolumn{1}{c|}{1} & \multicolumn{1}{c|}{0} & \multicolumn{1}{c|}{} & \multicolumn{1}{c|}{20.26 $\pm$ 9.47} & \multicolumn{1}{c|}{{\color[HTML]{002060} -15.65 (-44\%)}} & \multicolumn{1}{c|}{{\color[HTML]{002060} +0.57$\pm$ 0.43}} & \multicolumn{1}{c|}{0.25} \\ \cline{3-4} \cline{6-9} 
\multicolumn{1}{|c|}{} & \multicolumn{1}{c|}{} & \multicolumn{1}{c|}{0} & \multicolumn{1}{c|}{1} & \multicolumn{1}{c|}{} & \multicolumn{1}{c|}{21.02 $\pm$ 7.51} & \multicolumn{1}{c|}{{\color[HTML]{002060} -14.90 (-39\%)}} & \multicolumn{1}{c|}{{\color[HTML]{C00000} -0.07$\pm$ 0.38}} & \multicolumn{1}{c|}{0.56} \\ \cline{3-4} \cline{6-9} 
\multicolumn{1}{|c|}{} & \multicolumn{1}{c|}{\multirow{-3}{*}{MonoDepth2}} & \multicolumn{1}{c|}{0.85} & \multicolumn{1}{c|}{0.15} & \multicolumn{1}{c|}{\multirow{-3}{*}{35.92 $\pm$ 9.93}} & \multicolumn{1}{c|}{\textbf{17.00} $\pm$ 8.08} & \multicolumn{1}{c|}{{\color[HTML]{002060} -18.92 (-52\%)}} & \multicolumn{1}{c|}{{\color[HTML]{002060} +0.47$\pm$ 0.44}} & \multicolumn{1}{c|}{0.38} \\ \hhline{~========} 
\multicolumn{1}{|c|}{} & \multicolumn{1}{c|}{} & \multicolumn{1}{c|}{1} & \multicolumn{1}{c|}{0} & \multicolumn{1}{c|}{} & \multicolumn{1}{c|}{\textbf{7.97} $\pm$ 5.02} & \multicolumn{1}{c|}{{\color[HTML]{002060} -15.76 (-68\%)}} & \multicolumn{1}{c|}{{\color[HTML]{002060} +0.67$\pm$ 0.19}} & \multicolumn{1}{c|}{0.06} \\ \cline{3-4} \cline{6-9} 
\multicolumn{1}{|c|}{} & \multicolumn{1}{c|}{} & \multicolumn{1}{c|}{0} & \multicolumn{1}{c|}{1} & \multicolumn{1}{c|}{} & \multicolumn{1}{c|}{13.71 $\pm$ 6.11} & \multicolumn{1}{c|}{{\color[HTML]{002060} -10.03 (-40\%)}} & \multicolumn{1}{c|}{{\color[HTML]{C00000} +0.10$\pm$ 0.34}} & \multicolumn{1}{c|}{0.56} \\ \cline{3-4} \cline{6-9} 
\multicolumn{1}{|c|}{} & \multicolumn{1}{c|}{\multirow{-3}{*}{MADNet}} & \multicolumn{1}{c|}{0.85} & \multicolumn{1}{c|}{0.15} & \multicolumn{1}{c|}{\multirow{-3}{*}{23.74 $\pm$   9.68}} & \multicolumn{1}{c|}{\textbf{7.97} $\pm$ 5.38} & \multicolumn{1}{c|}{{\color[HTML]{002060} -15.77 (-68\%)}} & \multicolumn{1}{c|}{{\color[HTML]{002060} +0.60$\pm$ 0.32}} & \multicolumn{1}{c|}{0.12} \\ \hhline{~========} 
\multicolumn{1}{|c|}{} & \multicolumn{1}{c|}{} & \multicolumn{1}{c|}{1} & \multicolumn{1}{c|}{0} & \multicolumn{1}{c|}{} & \multicolumn{1}{c|}{\textbf{2.96} $\pm$ 1.77} & \multicolumn{1}{c|}{{\color[HTML]{C00000} +0.83 (+41\%)}} & \multicolumn{1}{c|}{{\color[HTML]{C00000} -0.22$\pm$ 0.81}} & \multicolumn{1}{c|}{0.56} \\ \cline{3-4} \cline{6-9} 
\multicolumn{1}{|c|}{} & \multicolumn{1}{c|}{} & \multicolumn{1}{c|}{0} & \multicolumn{1}{c|}{1} & \multicolumn{1}{c|}{} & \multicolumn{1}{c|}{10.27 $\pm$ 5.46} & \multicolumn{1}{c|}{{\color[HTML]{C00000} +8.14 (+420\%)}} & \multicolumn{1}{c|}{{\color[HTML]{C00000} -0.97$\pm$ 0.05}} & \multicolumn{1}{c|}{1.00} \\ \cline{3-4} \cline{6-9} 
\multicolumn{1}{|c|}{} & \multicolumn{1}{c|}{\multirow{-3}{*}{HSM}} & \multicolumn{1}{c|}{0.85} & \multicolumn{1}{c|}{0.15} & \multicolumn{1}{c|}{\multirow{-3}{*}{2.13 $\pm$ 0.73}} & \multicolumn{1}{c|}{3.58 $\pm$ 2.17} & \multicolumn{1}{c|}{{\color[HTML]{C00000} +1.44 (+70\%)}} & \multicolumn{1}{c|}{{\color[HTML]{C00000} -0.49$\pm$ 0.58}} & \multicolumn{1}{c|}{0.75} \\ \hhline{~========} 
\multicolumn{1}{|c|}{} & \multicolumn{1}{c|}{} & \multicolumn{1}{c|}{1} & \multicolumn{1}{c|}{0} & \multicolumn{1}{c|}{} & \multicolumn{1}{c|}{\textbf{2.26}$\pm$ 0.73} & \multicolumn{1}{c|}{{\color[HTML]{C00000} +0.03 (+6\%)}} & \multicolumn{1}{c|}{{\color[HTML]{C00000} -0.86$\pm$ 0.17}} & \multicolumn{1}{c|}{1.00} \\ \cline{3-4} \cline{6-9} 
\multicolumn{1}{|c|}{} & \multicolumn{1}{c|}{} & \multicolumn{1}{c|}{0} & \multicolumn{1}{c|}{1} & \multicolumn{1}{c|}{} & \multicolumn{1}{c|}{7.24$\pm$ 2.57} & \multicolumn{1}{c|}{{\color[HTML]{C00000} +4.97 (+251\%)}} & \multicolumn{1}{c|}{{\color[HTML]{C00000} -0.90$\pm$ 0.18}} & \multicolumn{1}{c|}{1.00} \\ \cline{3-4} \cline{6-9} 
\multicolumn{1}{|c|}{} & \multicolumn{1}{c|}{\multirow{-3}{*}{GwcNet}} & \multicolumn{1}{c|}{0.85} & \multicolumn{1}{c|}{0.15} & \multicolumn{1}{c|}{\multirow{-3}{*}{2.22$\pm$ 0.85}} & \multicolumn{1}{c|}{2.76$\pm$ 0.80} & \multicolumn{1}{c|}{{\color[HTML]{C00000} +0.54 (+32\%)}} & \multicolumn{1}{c|}{{\color[HTML]{C00000} -0.96$\pm$ 0.05}} & \multicolumn{1}{c|}{1.00} \\ \hhline{~========} 
\multicolumn{1}{|c|}{} & \multicolumn{1}{c|}{} & \multicolumn{1}{c|}{1} & \multicolumn{1}{c|}{0} & \multicolumn{1}{c|}{} & \multicolumn{1}{c|}{\textbf{3.20} $\pm$ 1.58} & \multicolumn{1}{c|}{{\color[HTML]{002060} -0.24 (-2\%)}} & \multicolumn{1}{c|}{{\color[HTML]{C00000} -0.26$\pm$ 0.55}} & \multicolumn{1}{c|}{0.69} \\ \cline{3-4} \cline{6-9} 
\multicolumn{1}{|c|}{} & \multicolumn{1}{c|}{} & \multicolumn{1}{c|}{0} & \multicolumn{1}{c|}{1} & \multicolumn{1}{c|}{} & \multicolumn{1}{c|}{9.49 $\pm$ 5.36} & \multicolumn{1}{c|}{{\color[HTML]{C00000} +6.05 (+197\%)}} & \multicolumn{1}{c|}{{\color[HTML]{C00000} -0.96$\pm$ 0.07}} & \multicolumn{1}{c|}{1.00} \\ \cline{3-4} \cline{6-9} 
\multicolumn{1}{|c|}{\multirow{-15}{*}{SERV-CT}} & \multicolumn{1}{c|}{\multirow{-3}{*}{STTR}} & \multicolumn{1}{c|}{0.85} & \multicolumn{1}{c|}{0.15} & \multicolumn{1}{c|}{\multirow{-3}{*}{3.44 $\pm$ 2.02}} & \multicolumn{1}{c|}{3.48 $\pm$ 1.83} & \multicolumn{1}{c|}{{\color[HTML]{C00000} +0.03 (+5\%)}} & \multicolumn{1}{c|}{{\color[HTML]{C00000} -0.39$\pm$ 0.53}} & \multicolumn{1}{c|}{0.81} \\ \hline
\multicolumn{1}{l}{} & \multicolumn{1}{l}{} & \multicolumn{1}{l}{} & \multicolumn{1}{l}{} &  &  & {\color[HTML]{C00000} } & {\color[HTML]{C00000} } &  \\ \hline
\multicolumn{1}{|c|}{} & \multicolumn{1}{c|}{} & \multicolumn{1}{c|}{1} & \multicolumn{1}{c|}{0} & \multicolumn{1}{c|}{} & \multicolumn{1}{c|}{11.24$\pm$ 5.48} & \multicolumn{1}{c|}{{\color[HTML]{002060} -3.39 (-24\%)}} & \multicolumn{1}{c|}{{\color[HTML]{002060} +0.41$\pm$ 0.87}} & \multicolumn{1}{c|}{0.30} \\ \cline{3-4} \cline{6-9} 
\multicolumn{1}{|c|}{} & \multicolumn{1}{c|}{} & \multicolumn{1}{c|}{0} & \multicolumn{1}{c|}{1} & \multicolumn{1}{c|}{} & \multicolumn{1}{c|}{\textbf{7.68}$\pm$ 2.83} & \multicolumn{1}{c|}{{\color[HTML]{002060} -6.15 (-43\%)}} & \multicolumn{1}{c|}{{\color[HTML]{002060} +0.79$\pm$ 0.53}} & \multicolumn{1}{c|}{0.10} \\ \cline{3-4} \cline{6-9} 
\multicolumn{1}{|c|}{} & \multicolumn{1}{c|}{\multirow{-3}{*}{MonoDepth2}} & \multicolumn{1}{c|}{0.85} & \multicolumn{1}{c|}{0.15} & \multicolumn{1}{c|}{\multirow{-3}{*}{14.54$\pm$   4.11}} & \multicolumn{1}{c|}{10.95$\pm$ 5.35} & \multicolumn{1}{c|}{{\color[HTML]{002060} -3.58 (-25\%)}} & \multicolumn{1}{c|}{{\color[HTML]{002060} +0.48$\pm$ 0.72}} & \multicolumn{1}{c|}{0.30} \\ \hhline{~========} 
\multicolumn{1}{|c|}{} & \multicolumn{1}{c|}{} & \multicolumn{1}{c|}{1} & \multicolumn{1}{c|}{0} & \multicolumn{1}{c|}{} & \multicolumn{1}{c|}{\textbf{17.26}$\pm$ 5.56} & \multicolumn{1}{c|}{{\color[HTML]{002060} -10.34 (-27\%)}} & \multicolumn{1}{c|}{{\color[HTML]{C00000} +0.16$\pm$ 0.50}} & \multicolumn{1}{c|}{0.50} \\ \cline{3-4} \cline{6-9} 
\multicolumn{1}{|c|}{} & \multicolumn{1}{c|}{} & \multicolumn{1}{c|}{0} & \multicolumn{1}{c|}{1} & \multicolumn{1}{c|}{} & \multicolumn{1}{c|}{22.12$\pm$ 10.10} & \multicolumn{1}{c|}{{\color[HTML]{002060} -5.55 (-13\%)}} & \multicolumn{1}{c|}{{\color[HTML]{C00000} -0.03$\pm$ 0.52}} & \multicolumn{1}{c|}{0.40} \\ \cline{3-4} \cline{6-9} 
\multicolumn{1}{|c|}{} & \multicolumn{1}{c|}{\multirow{-3}{*}{MADNet}} & \multicolumn{1}{c|}{0.85} & \multicolumn{1}{c|}{0.15} & \multicolumn{1}{c|}{\multirow{-3}{*}{27.49$\pm$ 13.09}} & \multicolumn{1}{c|}{17.54$\pm$ 6.23} & \multicolumn{1}{c|}{{\color[HTML]{002060} -9.95 (-25\%)}} & \multicolumn{1}{c|}{{\color[HTML]{C00000} +0.16$\pm$ 0.52}} & \multicolumn{1}{c|}{0.40} \\ \hhline{~========} 
\multicolumn{1}{|c|}{} & \multicolumn{1}{c|}{} & \multicolumn{1}{c|}{1} & \multicolumn{1}{c|}{0} & \multicolumn{1}{c|}{} & \multicolumn{1}{c|}{1.28$\pm$ 0.58} & \multicolumn{1}{c|}{{\color[HTML]{002060} -0.67 (-33\%)}} & \multicolumn{1}{c|}{{\color[HTML]{002060} +0.83$\pm$ 0.50}} & \multicolumn{1}{c|}{0.10} \\ \cline{3-4} \cline{6-9} 
\multicolumn{1}{|c|}{} & \multicolumn{1}{c|}{} & \multicolumn{1}{c|}{0} & \multicolumn{1}{c|}{1} & \multicolumn{1}{c|}{} & \multicolumn{1}{c|}{1.41$\pm$ 0.60} & \multicolumn{1}{c|}{{\color[HTML]{002060} -0.54 (-27\%)}} & \multicolumn{1}{c|}{{\color[HTML]{002060} +0.43$\pm$ 0.61}} & \multicolumn{1}{c|}{0.30} \\ \cline{3-4} \cline{6-9} 
\multicolumn{1}{|c|}{} & \multicolumn{1}{c|}{\multirow{-3}{*}{HSM}} & \multicolumn{1}{c|}{0.85} & \multicolumn{1}{c|}{0.15} & \multicolumn{1}{c|}{\multirow{-3}{*}{2.05$\pm$   0.90}} & \multicolumn{1}{c|}{\textbf{1.27}$\pm$ 0.58} & \multicolumn{1}{c|}{{\color[HTML]{002060} -0.68 (-34\%)}} & \multicolumn{1}{c|}{{\color[HTML]{C00000} +0.16$\pm$ 0.52}} & \multicolumn{1}{c|}{0.10} \\ \hhline{~========} 
\multicolumn{1}{|c|}{} & \multicolumn{1}{c|}{} & \multicolumn{1}{c|}{1} & \multicolumn{1}{c|}{0} & \multicolumn{1}{c|}{} & \multicolumn{1}{c|}{0.84$\pm$ 0.29} & \multicolumn{1}{c|}{{\color[HTML]{002060} -0.16 (-16\%)}} & \multicolumn{1}{c|}{{\color[HTML]{002060} +0.57$\pm$ 0.61}} & \multicolumn{1}{c|}{0.10} \\ \cline{3-4} \cline{6-9} 
\multicolumn{1}{|c|}{} & \multicolumn{1}{c|}{} & \multicolumn{1}{c|}{0} & \multicolumn{1}{c|}{1} & \multicolumn{1}{c|}{} & \multicolumn{1}{c|}{1.05$\pm$ 0.38} & \multicolumn{1}{c|}{{\color[HTML]{C00000} +0.05 (+5\%)}} & \multicolumn{1}{c|}{{\color[HTML]{C00000} -0.55$\pm$ 0.52}} & \multicolumn{1}{c|}{0.80} \\ \cline{3-4} \cline{6-9} 
\multicolumn{1}{|c|}{} & \multicolumn{1}{c|}{\multirow{-3}{*}{GwcNet}} & \multicolumn{1}{c|}{0.85} & \multicolumn{1}{c|}{0.15} & \multicolumn{1}{c|}{\multirow{-3}{*}{1.00$\pm$ 0.37}} & \multicolumn{1}{c|}{\textbf{0.83}$\pm$ 0.26} & \multicolumn{1}{c|}{{\color[HTML]{002060} -0.17 (-17\%)}} & \multicolumn{1}{c|}{{\color[HTML]{C00000} +0.19$\pm$ 0.59}} & \multicolumn{1}{c|}{0.60} \\ \hhline{~========} 
\multicolumn{1}{|c|}{} & \multicolumn{1}{c|}{} & \multicolumn{1}{c|}{1} & \multicolumn{1}{c|}{0} & \multicolumn{1}{c|}{} & \multicolumn{1}{c|}{1.69$\pm$ 1.65} & \multicolumn{1}{c|}{{\color[HTML]{C00000} +0.03 (+3\%)}} & \multicolumn{1}{c|}{{\color[HTML]{C00000} -0.61$\pm$ 0.54}} & \multicolumn{1}{c|}{0.80} \\ \cline{3-4} \cline{6-9} 
\multicolumn{1}{|c|}{} & \multicolumn{1}{c|}{} & \multicolumn{1}{c|}{0} & \multicolumn{1}{c|}{1} & \multicolumn{1}{c|}{} & \multicolumn{1}{c|}{1.93$\pm$ 1.33} & \multicolumn{1}{c|}{{\color[HTML]{C00000} +0.27 (+37\%)}} & \multicolumn{1}{c|}{{\color[HTML]{C00000} -0.36$\pm$ 0.72}} & \multicolumn{1}{c|}{0.90} \\ \cline{3-4} \cline{6-9} 
\multicolumn{1}{|c|}{\multirow{-15}{*}{Middlebury2014}} & \multicolumn{1}{c|}{\multirow{-3}{*}{STTR}} & \multicolumn{1}{c|}{0.85} & \multicolumn{1}{c|}{0.15} & \multicolumn{1}{c|}{\multirow{-3}{*}{1.66$\pm$ 2.60}} & \multicolumn{1}{c|}{\textbf{1.34}$\pm$ 1.23} & \multicolumn{1}{c|}{{\color[HTML]{002060} -0.31 (-19\%)}} & \multicolumn{1}{c|}{{\color[HTML]{C00000} -0.57$\pm$ 0.65}} & \multicolumn{1}{c|}{0.87} \\ \hline
\end{tabular}%
}
\end{table}

\section{Effect of Kernel Size and Type in SSIM}
\begin{figure}[htpb]
    \centering
    \subfloat[Linear Kernel (pixels are equally weighted)]{\includegraphics[height=0.27\linewidth]{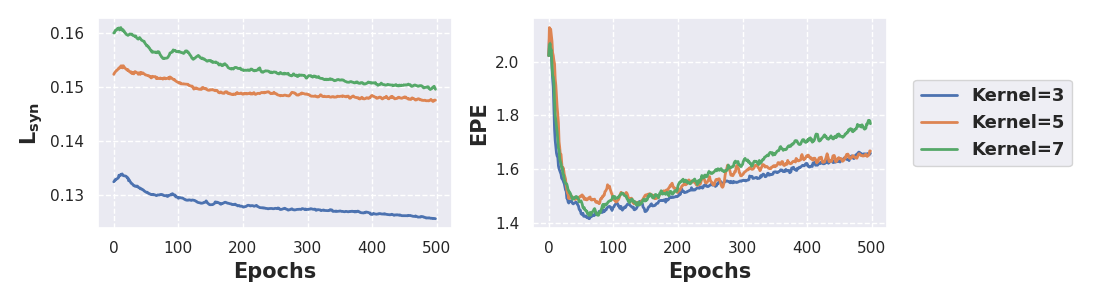}}
    
    \subfloat[Gaussian Kernel]{\includegraphics[height=0.27\linewidth]{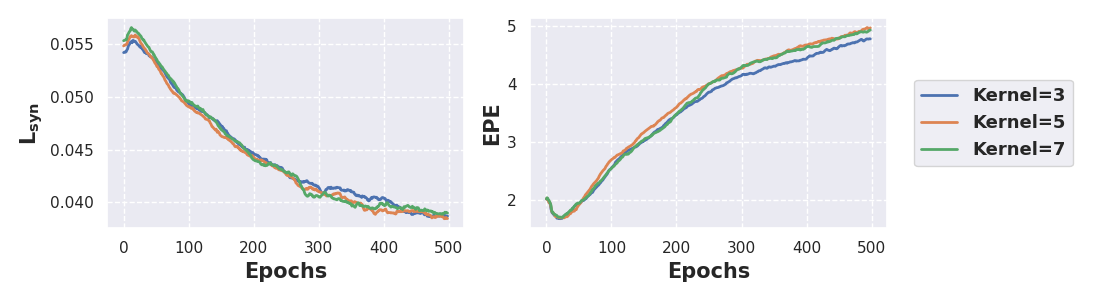}}
    \caption{Comparison of kernel size and kernel type for \ssim~computation.}
    \label{fig:kerenl}
\end{figure}

As mentioned in Section 3.2, there can be variations of the \ssim. One parameter is the size of the local patch for computing $\mu,\sigma$. The other parameter is the type of kernel used for computing $\mu,\sigma$, one can use either linear kernels (where each pixels in the local patch are weighted equally) or Gaussian kernels. We conduct experiments to examine the effects of kernel size and kernel type for \syn. We demonstrate that these parameters do not inherently change the training behavior as shown in \autoref{fig:kerenl}.

\section{Effect of Optimizer}
\begin{figure}[htpb]
    \centering
    \includegraphics[height=0.27\linewidth]{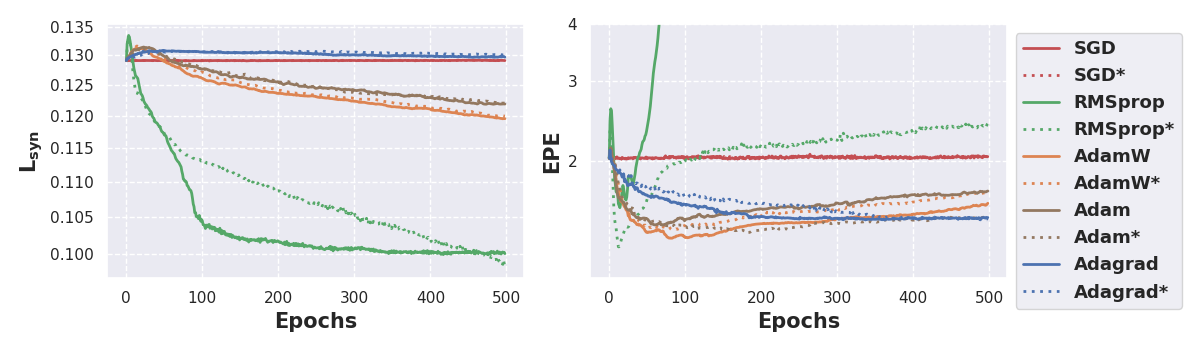}
    \caption{Ablation of different optimizers. Asterisk indicates presence of weight decay.}
    \label{fig:optimizer}
\end{figure}

To eliminate the possibility that the specific choice of optimizer causes the disagreement, we conduct an ablation study for different optimizers using STTR, as STTR exhibits the worst divergence, including AdamW~\citep{loshchilov2017decoupled}, Adam~\citep{kingma2014adam}, Adagrad~\citep{duchi2011adaptive}, RMSprop~\citep{graves2013generating} and SGD with momentum~\citep{sutskever2013importance}. All experiments are performed twice, with and without weight decay. The momentum is set to 0.9 where applicable. All experiments are performed with multi-scale and smoothness regularization. As shown in \autoref{fig:optimizer}, SGD suffers from gradient locality of $L_{syn}$, and does not make sufficient updates to the model. Adam, AdamW and RMSprop all lead to divergence between \syn~and depth prediction error after a certain point, especially RMSprop which minimizes \syn~the most. The only exception amongst the adaptive optimizers is Adagrad, which stops updating after a finite number of iterations due to gradient accumulation. This prevents \syn~from decreasing and thus avoids the inconsistency between \syn~and EPE, which can be a temporary alternative. However, because there is no good control over when the gradient magnitude will diminish with Adagrad, we feel that it is inadequate in practice.

\section{Synthesis Loss Beyond Intensities}
\begin{figure}[htpb]
    \centering
    \includegraphics[height=0.35\linewidth]{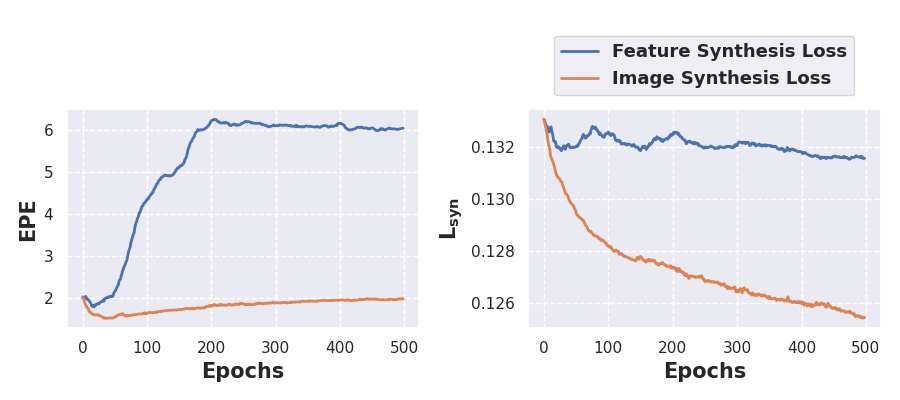}
    \caption{Ablation of \syn~on features and image intensities.}
    \label{fig:feature_syn}
\end{figure}

An alternative to using image intensities for $L_{syn}$ would be to use features extracted by the network. However, if a network already has the ability to distinguish each pixel and match them properly, the network should already predict the perfect disparity. Furthermore, we hypothesize that this learning paradigm introduces instability as the network can freely alter the feature representation compared to computing \syn~on image intensities. Regardless, we conduct a single experiment using features extracted by the network to compute $L_{syn}$ to verify the hypothesis. The experiment is conducted on STTR with both multi-scale and smoothness regularization. As shown in \autoref{fig:feature_syn}, due to a lack of constraints, the EPE increases much more faster compared to computing \syn~using image intensities.

\section{Survey of Papers Using Image synthesis Loss}
Following PRISMA guidelines~\citep{moher2009preferred}, we conducted a literature review on self-supervised depth estimation using \syn. Since the primary objective of this review is to estimate the scope of image synthesis-based self-supervision, we surveyed the arXiv database with the following search keywords:
\begin{itemize}
    \item For monocular depth estimation - (self-supervised $\lor$ unsupervised) $\land$ (depth $\lor$ disparity) $\land$ (estimation $\lor$ prediction); 
    \item For stereo depth estimation - (self-supervised $\lor$ unsupervised $\lor$ adapt) $\land$ (depth $\lor$ disparity) $\land$ (stereo). 
\end{itemize}
This search retrieved a total of 447 papers that were then screened further based on the following inclusion criteria:
\begin{itemize}
    \item Must be monocular or stereo depth estimation from images;
    \item Must use \syn~loss or other combination of \ssim~and \lone;
    \item Must be original primary investigation (i.e. not a review paper);
    \item Language must be English;
    \item Published 2015 and up to June 1st, 2021.
\end{itemize}
After screening, 190 papers were included for full text review. A total of 63 papers were subsequently rejected after full text review, leaving 127 papers that fit the inclusion criteria. Among the selected papers, 114 papers use smoothness regularization, 67 papers use multi-scale regularization and 69 papers handle occlusion explicitly.

\scriptsize
\begin{landscape}
\begin{longtable}[c]{|p{1.7cm}|p{10cm}|p{2.5cm}|p{0.8cm}|p{0.8cm}|p{1.1cm}|p{0.8cm}|p{1.6cm}|}
\caption{Summary of literature review on self-supervised depth estimation using image synthesis. M/S: Depth from monocular or stereo images. SMTH: Smoothness regularization. MS: Multi-scale regularization. Occlusion: Occlusion handling.}
\label{tab:lit-review}\\
\hline
\rotatebox[origin=c]{0}{\textbf{Paper ID}} & \rotatebox[origin=c]{0}{\textbf{Title}} & \rotatebox[origin=c]{0}{\textbf{Author Name}} & \rotatebox[origin=c]{0}{\textbf{Year}} & \rotatebox[origin=c]{0}{\textbf{M/S}} & \rotatebox[origin=c]{0}{\textbf{SMTH}} & \rotatebox[origin=c]{0}{\textbf{MS}} & \rotatebox[origin=c]{0}{\textbf{Occlusion}} \\ \hline
\endfirsthead
\endhead
1 & Improved Point Transformation Methods For Self-Supervised Depth Prediction & Ziwen & 2021 & S & x & x & \checkmark \\ \hline
2 & Learning Depth via Leveraging Semantics: Self-supervised Monocular DepthEstimation with Both Implicit and Explicit Semantic Guidance & Li & 2021 & M & \checkmark & \checkmark & x \\ \hline
3 & Learning Monocular Depth in Dynamic Scenes viaInstance-Aware Projection Consistency & Lee & 2021 & M & \checkmark & \checkmark & x \\ \hline
4 & Self-supervised monocular depth estimation from oblique UAV videos & Madhuanand & 2020 & M & \checkmark & x & \checkmark \\ \hline
5 & Semantic-Guided Representation Enhancement for Self-supervised Monocular Trained Depth Estimation & Li & 2020 & M & \checkmark & \checkmark & x \\ \hline
6 & HR-Depth : High Resolution Self-Supervised Monocular Depth Estimation & Lyu & 2020 & M & \checkmark & \checkmark & \checkmark \\ \hline
7 & Variational Monocular Depth Estimation for Reliability Prediction & Hirose & 2020 & M & \checkmark & x & x \\ \hline
8 & Attentional Separation-and-Aggregation Network for Self-supervised Depth-Pose Learning in Dynamic Scenes & Gao & 2020 & M & \checkmark & x & \checkmark \\ \hline
9 & Unsupervised Monocular Depth Learning with Integrated Intrinsics and Spatio-Temporal Constraints & Chen & 2020 & M & \checkmark & x & x \\ \hline
10 & Unsupervised Deep Persistent Monocular Visual Odometry and Depth Estimation in Extreme Environments & Almalioglu & 2020 & M & \checkmark & x & \checkmark \\ \hline
11 & Unsupervised Monocular Depth Learning in Dynamic Scenes & Li & 2020 & M & \checkmark & x & \checkmark \\ \hline
12 & Geometry-based Occlusion-Aware Unsupervised Stereo Matching for Autonomous Driving & Peng & 2020 & S & \checkmark & x & \checkmark \\ \hline
13 & Unsupervised Learning of Depth and Ego-Motion from Cylindrical Panoramic Video with Applications for Virtual Reality & Sharma & 2020 & M & \checkmark & \checkmark & x \\ \hline
14 & SAFENet: Self-Supervised Monocular Depth Estimation with Semantic-Aware Feature Extraction & Choi & 2020 & M & \checkmark & x & x \\ \hline
15 & Calibrating Self-supervised Monocular Depth Estimation & McCraith & 2020 & M & \checkmark & x & x \\ \hline
16 & Cascade Network for Self-Supervised Monocular Depth Estimation & Chai & 2020 & M & \checkmark & \checkmark & x \\ \hline
17 & Self-Supervised Learning for Monocular Depth Estimation from Aerial Imagery & Hermann & 2020 & M & \checkmark & \checkmark & \checkmark \\ \hline
18 & Reversing the cycle: self-supervised deep stereo through enhanced monocular distillation & Aleotti & 2020 & S & \checkmark & \checkmark & \checkmark \\ \hline
19 & Neural Ray Surfaces for Self-Supervised Learning of Depth and Ego-motion & Vasiljevic & 2020 & M & \checkmark & x & x \\ \hline
20 & SynDistNet: Self-Supervised Monocular Fisheye Camera Distance Estimation Synergized with Semantic Segmentation for Autonomous Driving & Kumar & 2020 & M & \checkmark & x & \checkmark \\ \hline
21 & S3Net: Semantic-Aware Self-supervised Depth Estimation with Monocular Videos and Synthetic Data & Cheng & 2020 & M & \checkmark & x & x \\ \hline
22 & P2Net: Patch-match and Plane-regularization for Unsupervised Indoor Depth Estimation & Yu & 2020 & M & \checkmark & x & x \\ \hline
23 & P2D: a self-supervised method for depth estimation from polarimetry & Blanchon & 2020 & M & \checkmark & x & \checkmark \\ \hline
24 & Self-Supervised Monocular Depth Estimation: Solving the Dynamic Object Problem by Semantic Guidance & Klingner & 2020 & M & \checkmark & \checkmark & x \\ \hline
25 & UnRectDepthNet: Self-Supervised Monocular Depth Estimation using a Generic Framework for Handling Common Camera Distortion Models & Kumar & 2020 & M & \checkmark & \checkmark & \checkmark \\ \hline
26 & Continual Adaptation for Deep Stereo & Poggii & 2020 & S & \checkmark & \checkmark & x \\ \hline
27 & Self-supervised Depth Estimation to Regularise Semantic Segmentation in Knee Arthroscopy & Liu & 2020 & M & x & \checkmark & x \\ \hline
28 & EndoSLAM Dataset and An Unsupervised Monocular Visual Odometry and Depth Estimation Approach for Endoscopic Videos: Endo-SfMLearner & Ozyoruk & 2020 & M & \checkmark & x & x \\ \hline
29 & MiniNet: An extremely lightweight convolutional neural network for real-time unsupervised monocular depth estimation & Liu & 2020 &  & \checkmark & \checkmark & \checkmark \\ \hline
30 & Increased-Range Unsupervised Monocular Depth Estimation & Imran & 2020 & M & \checkmark & \checkmark & \checkmark \\ \hline
31 & Consistency Guided Scene Flow Estimation & Chen & 2020 & S & \checkmark & x & \checkmark \\ \hline
32 & Self-Supervised Joint Learning Framework of Depth Estimation via Implicit Cues & Wang & 2020 & M, S & \checkmark & \checkmark & \checkmark \\ \hline
33 & Semantics-Driven Unsupervised Learning for Monocular Depth and Ego-Motion Estimation & Wei & 2020 & M & \checkmark & x & \checkmark \\ \hline
34 & Unsupervised Depth Learning in Challenging Indoor Video: Weak Rectification to Rescue & Biang & 2020 & M & \checkmark & x & \checkmark \\ \hline
35 & Self-Attention Dense Depth Estimation Network for Unrectified Video Sequences & Mathew & 2020 & M & \checkmark & x & x \\ \hline
36 & Deep feature fusion for self-supervised monocular depth prediction & Kaushik & 2020 & M & \checkmark & \checkmark & \checkmark \\ \hline
37 & Self-Supervised Human Depth Estimation from Monocular Videos & Tan & 2020 & M & \checkmark & x & \checkmark \\ \hline
38 & Self-Supervised Attention Learning for Depth and Ego-motion Estimation & Sadek & 2020 & M & \checkmark & x & x \\ \hline
39 & Pseudo RGB-D for Self-Improving Monocular SLAM and Depth Prediction & Tiwari & 2020 & M & \checkmark & x & x \\ \hline
40 & RealMonoDepth: Self-Supervised Monocular Depth Estimation for General Scenes & Ocal & 2020 & M & \checkmark & \checkmark & \checkmark \\ \hline
41 & Masked GANs for Unsupervised Depth and Pose Prediction with Scale Consistency & Zhao & 2020 & M & \checkmark & x & \checkmark \\ \hline
42 & Self-Supervised Monocular Scene Flow Estimation & Hur & 2020 & M & \checkmark & x & \checkmark \\ \hline
43 & Distilled Semantics for Comprehensive Scene Understanding from Videos & Tosi & 2020 & M & \checkmark & x & \checkmark \\ \hline
44 & Self-supervised Monocular Trained Depth Estimation using Self-attention and Discrete Disparity Volume & Johnston & 2020 & M & \checkmark & \checkmark & \checkmark \\ \hline
45 & DeFeat-Net: General Monocular Depth via Simultaneous Unsupervised Representation Learning & Spencer & 2020 & M & \checkmark & x & \checkmark \\ \hline
46 & DiPE: Deeper into Photometric Errors for Unsupervised Learning of Depth and Ego-motion from Monocular Videos & Jiang & 2020 & M & \checkmark & \checkmark & \checkmark \\ \hline
47 & Unsupervised Learning of Depth, Optical Flow and Pose with Occlusion from 3D Geometry & Wang & 2020 & M & \checkmark & x & \checkmark \\ \hline
48 & Semantically-Guided Representation Learning for Self-Supervised Monocular Depth & Guiziliini & 2020 & M & \checkmark & \checkmark & \checkmark \\ \hline
49 & Single Image Depth Estimation Trained via Depth from Defocus Cues & Gur & 2020 & M & \checkmark & x & x \\ \hline
50 & Don’t Forget The Past: Recurrent Depth Estimation from Monocular Video & Patil & 2020 & M & \checkmark & \checkmark & \checkmark \\ \hline
51 & Self-supervised Object Motion and Depth Estimation from Video & Dai & 2020 & M & \checkmark & \checkmark & x \\ \hline
52 & Edge-Guided Occlusion Fading Reduction for a Light-Weighted Self-Supervised Monocular Depth Estimation & Peng & 2019 & M & \checkmark & x & \checkmark \\ \hline
53 & Unsupervised Monocular Depth Prediction for Indoor Continuous Video Streams & Feng & 2019 & M & \checkmark & \checkmark & \checkmark \\ \hline
54 & Unsupervised High-Resolution Depth Learning From Videos With Dual Networks & Zhou & 2019 & M & \checkmark & \checkmark & \checkmark \\ \hline
55 & FisheyeDistanceNet: Self-Supervised Scale-Aware Distance Estimation using Monocular Fisheye Camera for Autonomous Driving & Kumar & 2020 & M & \checkmark & x & \checkmark \\ \hline
56 & Self-Supervised Learning of Depth and Ego-motion with Differentiable Bundle Adjustment & Shi & 2019 & M & \checkmark & \checkmark & x \\ \hline
57 & Spherical View Synthesis for Self-Supervised 360o Depth Estimation & Zioulis & 2019 & M & \checkmark & x & x \\ \hline
58 & Progressive Fusion for Unsupervised Binocular Depth Estimation using Cycled Networks & Pilzer & 2019 & S & x & \checkmark & x \\ \hline
59 & Learning Residual Flow as Dynamic Motion from Stereo Videos & Lee & 2019 & S & \checkmark & \checkmark & \checkmark \\ \hline
60 & Unsupervised Domain Adaptation for Depth Prediction from Images & Tonioni & 2019 & M & \checkmark & x & \checkmark \\ \hline
61 & MVS2: Deep Unsupervised Multi-view Stereo with Multi-View Symmetry & Dai & 2019 & S & \checkmark & x & \checkmark \\ \hline
62 & Improving Self-Supervised Single View Depth Estimation by Masking Occlusion & Schellevis & 2019 & M & \checkmark & \checkmark & \checkmark \\ \hline
63 & Unsupervised Scale-consistent Depth and Ego-motion Learning from Monocular Video & Bian & 2019 & M & \checkmark & x & \checkmark \\ \hline
64 & Self-Supervised Learning With Geometric Constraints in Monocular Video: Connecting Flow, Depth, and Camera & Chen & 2019 & M & \checkmark & \checkmark & x \\ \hline
65 & Non-destructive three-dimensional measurement of hand vein based on self-supervised network & Chen & 2019 & S & \checkmark & x & x \\ \hline
66 & Bridging Stereo Matching and Optical Flow via Spatiotemporal Correspondence & Lai & 2019 & S & \checkmark & \checkmark & \checkmark \\ \hline
67 & Unsupervised Depth Completion from Visual Inertial Odometry & Wong & 2020 & M & \checkmark & x & x \\ \hline
68 & Semi-Supervised Monocular Depth Estimation with Left-Right Consistency Using Deep Neural Network & Amiri & 2019 & M & \checkmark & \checkmark & x \\ \hline
69 & Learning Unsupervised Multi-View Stereopsis via Robust Photometric Consistency & Khot & 2019 & S & \checkmark & x & \checkmark \\ \hline
70 & 3D Packing for Self-Supervised Monocular Depth Estimation & Guizilini & 2020 & M & \checkmark & \checkmark & \checkmark \\ \hline
71 & Learn Stereo, Infer Mono: Siamese Networks for Self-Supervised, Monocular, Depth Estimation & Goldman & 2019 & M, S & \checkmark & \checkmark & \checkmark \\ \hline
72 & Recurrent Neural Network for (Un-)supervised Learning of Monocular Video Visual Odometry and Depth & Wang & 2019 & M & \checkmark & \checkmark & \checkmark \\ \hline
73 & Learning monocular depth estimation infusing traditional stereo knowledge & Tosi & 2019 & M & \checkmark & \checkmark & \checkmark \\ \hline
74 & Learning to Adapt for Stereo & Tonioni & 2019 & S & x & x & \checkmark \\ \hline
75 & Geometry-Aware Symmetric Domain Adaptation for Monocular Depth Estimation & Zhao & 2019 & M & \checkmark & x & x \\ \hline
76 & A Novel Monocular Disparity Estimation Network with Domain Transformation and Ambiguity Learning & Bello & 2019 & M & \checkmark & \checkmark & \checkmark \\ \hline
77 & Refine and Distill: Exploiting Cycle-Inconsistency and Knowledge Distillation for Unsupervised Monocular Depth Estimation & Pilzer & 2019 & M & x & \checkmark & x \\ \hline
78 & Unsupervised Cross-spectral Stereo Matching by Learning to Synthesize & Liang & 2019 & S & \checkmark & \checkmark & x \\ \hline
79 & Region Deformer Networks for Unsupervised Depth Estimation from Unconstrained Monocular Videos & Xu & 2019 & M & \checkmark & \checkmark & \checkmark \\ \hline
80 & Unsupervised monocular stereo matching & Zhang & 2018 & M & \checkmark & \checkmark & x \\ \hline
81 & Unsupervised Learning of Monocular Depth Estimation with Bundle Adjustment, Super-Resolution and Clip Loss & Zhou & 2018 & M & \checkmark & \checkmark & \checkmark \\ \hline
82 & Joint Unsupervised Learning of Optical Flow and Depth by Watching Stereo Videos & Wang & 2018 & S & \checkmark & x & \checkmark \\ \hline
83 & SuperDepth: Self-Supervised, Super-Resolved Monocular Depth Estimation & Pillai & 2018 & M & \checkmark & \checkmark & \checkmark \\ \hline
84 & DispSegNet: Leveraging Semantics for End-to-End Learning of Disparity Estimation from Stereo Imagery & Zhang & 2019 & S & \checkmark & x & \checkmark \\ \hline
85 & Learning structure-from-motion from motion & Pinard & 2018 & M & \checkmark & \checkmark & \checkmark \\ \hline
86 & DF-Net: Unsupervised Joint Learning of Depth and Flow using Cross-Task Consistency & Zhou & 2018 & M & \checkmark & \checkmark & \checkmark \\ \hline
87 & A Deeper Insight into the UnDEMoN: Unsupervised Deep Network for Depth and Ego-Motion Estimation & Babu V & 2018 & M & \checkmark & \checkmark & x \\ \hline
88 & Learning monocular depth by distilling cross-domain stereo networks & Guo & 2018 & M & \checkmark & \checkmark & \checkmark \\ \hline
89 & Learning monocular depth estimation with unsupervised trinocular assumptions & Poggii & 2018 & M & \checkmark & \checkmark & \checkmark \\ \hline
90 & Towards real-time unsupervised monocular depth estimation on CPU & Poggii & 2018 & M & \checkmark & \checkmark & x \\ \hline
91 & Every Pixel Counts: Unsupervised Geometry Learning with Holistic 3D Motion Understanding & Yang & 2018 & M & \checkmark & \checkmark & \checkmark \\ \hline
92 & Digging Into Self-Supervised Monocular Depth Estimation & Godard & 2019 & M & \checkmark & \checkmark & \checkmark \\ \hline
93 & Competitive Collaboration: Joint Unsupervised Learning of Depth, Camera Motion, Optical Flow and Motion Segmentation & Ranjan & 2019 & M & \checkmark & \checkmark & \checkmark \\ \hline
94 & Position Estimation of Camera Based on Unsupervised Learning & Wu & 2018 & M & x & x & x \\ \hline
95 & Dual CNN Models for Unsupervised Monocular Depth Estimation & Repala & 2019 & M & \checkmark & \checkmark & x \\ \hline
96 & On the importance of Stereo for Accurate Depth Estimation: An Efficient Semi-Supervised Deep Neural Network Approach & Smolyanskiy & 2020 & S & \checkmark & x & x \\ \hline
97 & Fusion of stereo and still monocular depth estimates in a self-supervised learning context & Martins & 2018 & M, S & x & x & x \\ \hline
98 & LEGO: Learning Edge with Geometry all at Once by Watching Videos & Yang & 2018 & M & \checkmark & \checkmark & x \\ \hline
99 & Self-Supervised Monocular Image Depth Learning and Confidence Estimation & Chen & 2018 & M & \checkmark & \checkmark & x \\ \hline
100 & Unsupervised Learning of Monocular Depth Estimation and Visual Odometry with Deep Feature Reconstruction & Zhan & 2018 & S & \checkmark & x & x \\ \hline
101 & Geonet: Unsupervised learning of dense depth, optical flow and camera pose & Yin & 2018 & M & \checkmark & \checkmark & \checkmark \\ \hline
102 & AdaDepth: Unsupervised Content Congruent Adaptation for Depth Estimation & Kundu & 2018 & M & x & x & x \\ \hline
103 & Unsupervised Odometry and Depth Learning for Endoscopic Capsule Robots & Turan & 2018 & M & \checkmark & \checkmark & \checkmark \\ \hline
104 & Unsupervised Learning of Depth and Ego-Motion from Monocular Video Using 3D Geometric Constraints & Mahjourian & 2018 & M & \checkmark & \checkmark & x \\ \hline
105 & Learning Depth from Monocular Videos using Direct Methods & Wang & 2017 & M & \checkmark & \checkmark & x \\ \hline
106 & Unsupervised Learning of Geometry with Edge-aware Depth-Normal Consistency & Yang & 2017 & M & x & \checkmark & x \\ \hline
107 & UnDeepVO: Monocular Visual Odometry through Unsupervised Deep Learning & Li & 2018 & M & x & x & x \\ \hline
108 & Multi-task Self-Supervised Visual Learning & Doersch & 2017 & M & x & x & x \\ \hline
109 & Self-Supervised Siamese Learning on Stereo Image Pairs for Depth Estimation in Robotic Surgery & Ye & 2017 & S & x & x & x \\ \hline
110 & Unsupervised Learning of Depth and Ego-Motion from Video & Zhou & 2017 & M & \checkmark & \checkmark & x \\ \hline
111 & SfM-Net: Learning of Structure and Motion from Video & Vijayanarasimhan & 2017 & M & \checkmark & x & x \\ \hline
112 & Unsupervised monocular depth estimation with left-right consistency & Godard & 2017 & M & \checkmark & \checkmark & \checkmark \\ \hline
113 & Unsupervised CNN for Single View Depth Estimation: Geometry to the Rescue & Garg & 2016 & S & \checkmark & \checkmark & x \\ \hline
114 & EffiScene: Efficient Per-Pixel Rigidity Inference for Unsupervised Joint Learning of Optical Flow, Depth, Camera Pose and Motion Segmentation & Jiao & 2020 & S & \checkmark & x & \checkmark \\ \hline
115 & Parallax Attention for Unsupervised Stereo Correspondence Learning & Wang & 2020 & S & \checkmark & \checkmark & \checkmark \\ \hline
116 & Self-Supervised Scale Recovery for Monocular Depth and Egomotion Estimation & Wagstaff & 2020 & M & \checkmark & x & x \\ \hline
117 & Monocular Depth Estimation with Self-supervised Instance Adaptation & McCraith & 2020 & M & \checkmark & x & \checkmark \\ \hline
118 & Toward Hierarchical Self-Supervised Monocular Absolute Depth Estimation for Autonomous Driving Applications & Xue & 2020 & M & \checkmark & \checkmark & x \\ \hline
119 & AdaStereo: A Simple and Efficient Approach for Adaptive Stereo Matching & Song & 2020 & S & \checkmark & x & \checkmark \\ \hline
120 & Enhancing self-supervised monocular depth estimation with traditional visual odometry & Andraghetti & 2019 & S & \checkmark & \checkmark & \checkmark \\ \hline
121 & LiStereo: Generate Dense Depth Maps from LIDAR and Stereo Imagery & Zhang & 2020 & S & \checkmark & x & x \\ \hline
122 & Online Adaptation through Meta-Learning for Stereo Depth Estimation & Zhang & 2019 & S & x & x & x \\ \hline
123 & Bilateral Cyclic Constraint and Adaptive Regularization for Unsupervised Monocular Depth Prediction & Wong & 2019 & M & \checkmark & x & \checkmark \\ \hline
124 & Real-time self-adaptive deep stereo & Tonioni & 2019 & S & \checkmark & \checkmark & x \\ \hline
125 & Geometry meets semantics for semi-supervised monocular depth estimation & Ramirez & 2018 & M & \checkmark & \checkmark & \checkmark \\ \hline
126 & Open-World Stereo Video Matching with Deep RNN & Zhong & 2018 & S & \checkmark & x & x \\ \hline
127 & Self-Supervised Learning for Stereo Matching with Self-Improving Ability & Zhong & 2017 & S & \checkmark & x & x \\ \hline
\end{longtable}
\end{landscape}

\normalsize

\end{document}